# HMM-based Indic Handwritten Word Recognition using Zone Segmentation

[a]Partha Pratim Roy*, [b]Ayan Kumar Bhunia, [b]Ayan Das, [c]Prasenjit Dey, [d]Umapada Pal

[a]Dept. of CSE, Indian Institute of Technology Roorkee, India
[b]Dept. of ECE, Institute of Engineering & Management, Kolkata, India
[c]Dept. of CSE, Institute of Engineering & Management, Kolkata, India
[d]CVPR Unit, Indian Statistical Institute, Kolkata, India
[a]email: proy.fcs@iitr.ac.in, TEL: +91-1332-284816

## Abstract

This paper presents a novel approach towards Indic handwritten word recognition using zone-wise information. Because of complex nature due to compound characters, modifiers, overlapping and touching, etc., character segmentation and recognition is a tedious job in Indic scripts (e.g. Devanagari, Bangla, Gurumukhi, and other similar scripts). To avoid character segmentation in such scripts, HMM-based sequence modeling has been used earlier in holistic way. This paper proposes an efficient word recognition framework by segmenting the handwritten word images horizontally into three zones (upper, middle and lower) and recognize the corresponding zones. The main aim of this zone segmentation approach is to reduce the number of distinct component classes compared to the total number of classes in Indic scripts. As a result, use of this zone segmentation approach enhances the recognition performance of the system. The components in middle zone where characters are mostly touching are recognized using HMM. After the recognition of middle zone, HMM based Viterbi forced alignment is applied to mark the left and right boundaries of the characters. Next, the residue components, if any, in upper and lower zones in their respective boundary are combined to achieve the final word level recognition. Water reservoir feature has been integrated in this framework to improve the zone segmentation and character alignment defects while segmentation. A novel sliding window-based feature, called Pyramid Histogram of Oriented Gradient (PHOG) is proposed for middle zone recognition. PHOG features has been compared with other existing features and found robust in Indic script recognition. An exhaustive experiment is performed on two Indic scripts namely, Bangla and Devanagari for the performance evaluation. From the experiment, it has been noted that proposed zone-wise recognition improves accuracy with respect to the traditional way of Indic word recognition.

# 1. Introduction

Although, the automatic recognition of printed text has achieved a great success rate, the performance of handwritten word recognition is not high. Handwritten word recognition has long been an active research area because of its vast potential applications. Some of its potential application areas are postal automation, bank cheque processing, automatic data entry, etc. The main hindrance behind the difficulties of making a handwritten recognition system is the huge variation in writing style and complex shapes of characters in words. There are many research works towards handwritten word recognition in Roman [1], Japanese/Chinese [2, 3] and Arabic scripts [4]. Although many investigations have been made towards the recognition of isolated handwritten characters and digits of Indian scripts [5], only a few pieces of work [6, 7] exist towards offline handwritten word recognition in Indian scripts.

Devanagari and Bangla are two most popular Indian scripts. Devanagari, script is used to write languages such as Sanskrit, Hindi, Nepali, Marathi, and many others. It is used by approximately 400 million people in northern India and it is the most widely used Indic script. Bangla is the second most popular language in India. Languages like Bangla, Assamese and Manipuri languages are written in Bangla script. About 200 million people of Eastern India and Bangladesh use Bangla script for communication. Also, Devanagari is the third most and Bangla is the fifth most popular language in the world [8]. Examples of Bangla and Devanagari handwritten document images are shown in Fig. 1.

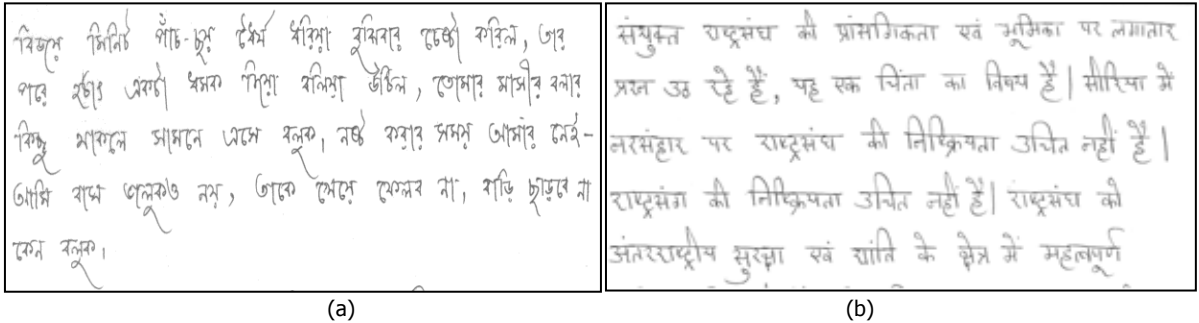

(a) (b)
**Fig.1: Examples of handwritten text document. (a) Bangla, (b) Devanagari**

The OCR involving printed Devanagari and Bangla scripts has been addressed in many pieces of research work [9, 10, 11]. Although a number of work has been investigated for isolated handwritten character and digit recognition in Indian script [5], only a few pieces of work exist towards handwritten word recognition in Indian script [6, 7]. Offline recognition of handwritten word of these scripts needs lot of research.

Most of the existing works in these two scripts are performed on segmenting the characters from words and then recognition. A number of works have been performed for character level segmentation in Devanagari [12] and Bangla [6]. It is reported that due to the presence of noise, touching, etc., the segmentation of characters from a word may often fail. Often characters may generate disjoint character components through preliminary segmentation process. Overlapping and touching characters, which frequently occur in Bangla writing style, create more hindrance in segmenting characters of the words.

In the past decades stochastic approaches such as Hidden Markov Models (HMMs) have been widely applied to perform word recognition task [13, 14] because of its effectiveness for modelling unconstrained character-string. This is mostly due to their ability to cope with non-linear distortions and incomplete information. Mainly two approaches namely segmentation-based approach [15] and holistic approach [16] are used for the word recognition purpose. In practice, a HMM can be employed to represent a whole word or, alternatively, sub-word units such as characters which can be concatenated to form general strings. Though HMMs-based techniques have been successfully used in handwriting recognition [17], only a few papers exist for Indian script recognition. One of the reasons could be the larger number of character classes in Indic scripts due to modifiers and compound characters.

Only a few pieces of work using HMMs are performed in Devanagari [18] and in Bangla [7] handwritten word recognition. Almost all these methods consider holistic approach of recognition as word-wise HMM model creation. In these approaches feature extraction was performed from the entire word and recognition was performed with the help of lexicon-based holistic word recognition. The main drawbacks of these holistic word-based HMMs models are that the recognition process is limited to a set of words only. Also, in this method, for each word a large number of training data is needed. An unknown word which was not trained by the models will not be recognized using these systems.

To overcome these drawbacks, HMMs are trained on sub-word units, such as characters, which can be concatenated to form general strings. Character based HMM models [13] have been successfully used for recognition of arbitrary set of words in English/Latin scripts. One of the advantages is that they allow recognizing unknown words from training data once the character models are trained. HMMs avoid the problem of pre-segmentation of words into characters so that the errors of pre-segmentation can be eliminated. Character alignment based techniques for HMM is also studied to reduce the error [3]. Note that such approach was not applied for offline word recognition of Indic scripts earlier.Though, this character based HMM models are popular in the literature of word recognition, the process may not be directly useful in Indic scripts, especially in Devanagari and Bangla. It is due to the fact that in such scripts, combination of vowels, modifiers and characters lead to a huge number of character classes.  Thus, sufficient data for each combination will be necessary for training the respective class models. To reduce such huge number of character classes we propose a zone-wise recognition approach where a word is segmented into 3 zones (upper, middle and lower zone). To have an idea about such character class reduction, let X, Y and Z be the number of character classes that may appear in upper, middle and lower zones, respectively. If we do not use zone-wise recognition then number of character classes will be XY+YZ (assuming all characters in the middle zone may be associated with all characters of upper and lower zones). Whereas if we use zone-wise segmentation, total number of character classes will be X+Y+Z instead of XY+YZ. Thus, there will be a huge reduction of characters when X, Y and Z are large. To have an idea, in Bangla we have about 280 characters (simple and compound together)[10] which may appear in middle zone and 4 modifiers in upper zone and 3 modifiers in lower zones. Thus if we do not use zone segmentation, we will have ideally 280×4+280×3= 1960 classes, whereas after zone segmentation we will have only 287 classes, Thus a reduction of 85.36% can be achieved. Based on this principle, recently we proposed a zone-wise recognition approach [19] and showed some preliminary results. This paper is an extension of the earlier paper including several additional contributions. The main contributions of this extended paper are the following: 1) integration of water reservoir concept for better zone segmentation in a word image, 2) efficient PHOG features developed to improve the performance of HMM based middle zone recognition, 3) the proposed framework has been generalized and tested for Bangla and Devanagari scripts recognition.

Overall organization of the rest of the paper is as follows. Section 2 describes some important properties and challenges in Bangla and Devanagari scripts. In Section 3, we describe the pre-processing tasks of Indic word images. The word-recognition framework using zone-wise segmentation results is explained in Section 4. Here details of zone segmentation, feature extraction and recognition approaches are discussed. We demonstrate the performance of the proposed approach in Section 5. Finally, conclusions are presented in Section 6.

## 2. Properties of Bangla and Devanagari Scripts

In Devanagari script, a total 49 basic characters exist, out of these 11 are vowels and 38 are consonants. The alphabet of the modern Bangla script consists of 11 vowels and 39 consonants. The basic characters of Bangla and Devanagari scripts are shown in Fig.2. It can be noted that most of the characters in Bangla and Devanagari scripts have a horizontal line (called *Matra/Shirorekha*) at the upper part and a baseline. When two or more characters sit side by side to form a word, these horizontal lines generally touch and generate a long line. Characters typically hang from the Matra when written. All Indic scripts run left to right, although some combining glyphs appear to the left of their base character for display. In both Bangla and Devanagari scripts a vowel following a consonant takes a modified shape and placed at the left, right, both left and right, or bottom of the consonant. These modified shapes are called *modified characters*. Examples of modified character are shown in Fig.2(i.b) and Fig. 2(ii.b) for Bangla and Devanagari scripts respectively. These modifiers add extra difficulty in the character segmentation procedure of Bangla and Devanagari scripts because of their topological position. A consonant or a vowel following a consonant sometimes takes a compound orthographic shape, which we call as *compound character*. For details about Bangla and Devanagari scripts, we refer [10].

(i)

(ii)

**Fig.2:Few examples of vowels, modifiers, consonants and conjuncts in (i) Bangla & (ii) Devanagari Script**

A Bangla or Devanagari word can be partitioned into three zones. The *upper-zone* ($Z_U$) denotes the portion above the *Matra*, the *middle zone* ($Z_M$) covers the portion between *Matra* and base-line, the *lower-zone* ($Z_L$) is the portion below base-line. Different zones in a Bangla word image are shown in Fig.3.

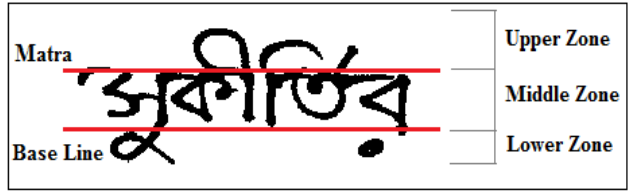

**Fig.3: Three zones of Bangla script – upper, middle and lower zone separated by Matra and base line**

## 2.1. Challenges in Bangla/Devanagari Word Recognition

As discussed above, recognition of Bangla and Devanagari script is not similar like Latin script due to the variation of character-modifiers presence in 3 zones: upper, middle and lower zones. When the consonant character, "ক"(appear only in middle zone)get combined with a vowel, the vowel forms a modifier which can appear either in middle zone (like "কা"), or in middle and upper zone (like "কি") according to the nature of vowel (as shown in Fig.2). Hence, the combinations of consonants and vowel make a large number of possible character combinations. Because of this, traditional HMM-based recognition systems (without zone segmentation)have to consider different character units for each combination separately, as the basic sliding window feature needs to capture the information in all zones for identifying the modifier properly. In Fig.4 (left column), it is graphically shown when a Bangla consonant character 'ক' is combined with 5 different vowels. In our proposed zone-segmentation based approach (right half of the Fig.4), it is possible to make these complex character shapes to model by few simple character units. For example, we can divide the shape কি into simple units ক, া, and ⌒.Similar is the case for other consonants of Bangla and Devanagari modifiers.

**Fig.4: example of character units reduction using zone segmentation**

Also, while writing Bangla/Devanagari characters, they suffer from distortions depending on the writing style of the person like other scripts. Often the *Matra* appears missing in handwritten words which also add challenges for developing a generalized recognition system. It is observed that due to the presence of noise, touching characters, etc., the segmentation of characters from a word may fail. Often characters may generate disjoint character components through preliminary segmentation process, which creates problem in recognition tasks. Proper classification and reunification of these components using segmentation are not easy to process. Overlapping and touching characters, which frequently occur in Bangla/Devanagari writing style, create more hindrance in recognizing characters of the words. Another problem is the "Slant and Skew" nature of handwritten word (see Fig.5). Due to non-uniform skew and slant in word images the recognition of words become more difficult. As mentioned earlier, the "Matra" stays in a horizontal line dividing the upper and mid-section of the word, which often fails to be so. Our recognition framework is designed to take care of these issues. These are discussed in the following section.

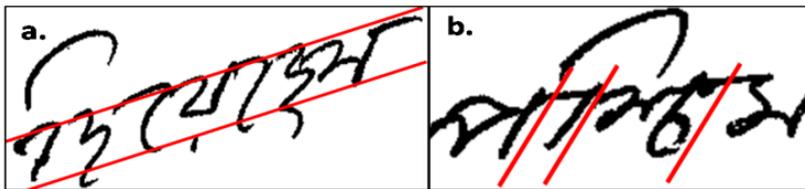

**Fig.5: (a) Skewed word. Red lines are the "Matra" and "Base line" respectively which are no longer horizontal. (b) Slanted word. Character segmentation lines are not vertical.**

## 3. Preprocessing

To extract the word image from the handwritten document a set of pre-processing tasks are followed. The offline document image is first converted into binary image using global histogram-based Otsu binarization method. The binary document is segmented into individual text lines using a line segmentation algorithm [20]. Here, some seed components of a line are obtained from smoothed text regions of document. The upper and lower boundary information of a text line is obtained from background regions using morphological functions. Next, foreground seed components and boundary information are used to segment the text-lines. Once lines are separated, Run Length Smoothing Algorithm (RLSA) [20] is next applied on each text line to get individual words as a component. A connected component labeling is applied to find the bounding box of the word patches in the line. Next, using the patch mask, the original word is considered from the binary image. The word images are next processed for skew and slant corrections. These are discussed in following subsection.

### 3.1. Skew and Slant Correction

In our framework "Water reservoir" concept has been applied for skew correction of non-horizontal words. This idea has been used earlier for various document image analysis techniques such as, script identification, line segmentation, etc. in Indian scripts [6, 10, 21]. In this concept, if water is poured from a side of a component, the cavity regions of the component where water will be stored are considered as reservoirs of the component. Because of touching through head-line in Bangla and Devanagari scripts, two consecutive characters in a word create large cavity regions (space) between them and hence we get large reservoirs [21] from the cavity regions. Water reservoir concept is not new but here we explore its application towards skew detection in Bangla and Devanagari scripts. Details of water reservoir and its different properties can be obtained in [21].

To use the concept of water reservoirs stated above to detect skew angle, we obtain the bottom reservoirs [21] of the word image (by pouring water from the bottom of the word). Then we filter out the reservoirs having lower heights, i.e. less than $3 \times S_w$, where $S_w$ is the average stroke width of the handwritten word image. Stroke width ($S_w$) is calculated as statistical mode of the run lengths of the word's foreground. A word image is, at first, scanned row-wise (horizontally) and then column-wise (vertically) to compute foreground pixel's run-lengths and their occurrence frequencies. Next, the statistical mode value of these run-lengths provides the estimated stroke width ($S_w$).

The local minima points (depth-points) from the valid reservoirs having heights more than $3 \times S_w$ are determined by traversing the contour. Let, $B$ be the set of all such points. Next, a first order degree polynomial (i.e. a straight line) using *Linear Regression* is computed using points of $B$. We have noted that the slope ($\theta$) of the calculated line provides a quantitative measurement of the skew nature of the handwritten word image. Thereafter the image is rotated in opposite direction by $\theta$ for skew correction. This process is illustrated in Fig.6(a).

The slant angle is next determined and corrected using the vertical projection histogram and Wigner–Ville distribution [22]. Using this projection histogram analysis, we find the height of the peaks in vertical projection analysis at an angle with an interval from -45° to +45° after doing shear transform. Next, the angle at which clear peaks and troughs are found is considered as slant angle. This is illustrated in Fig.6(b).

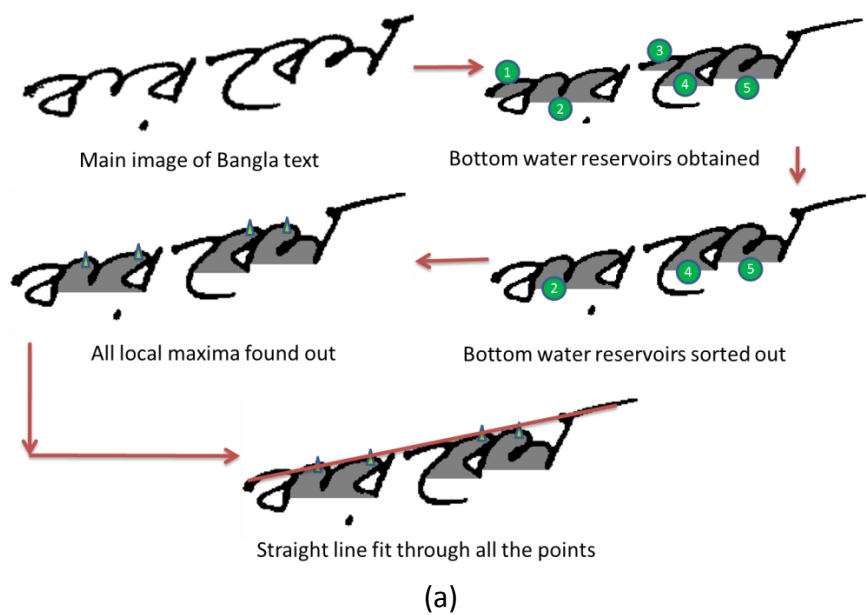

(a)

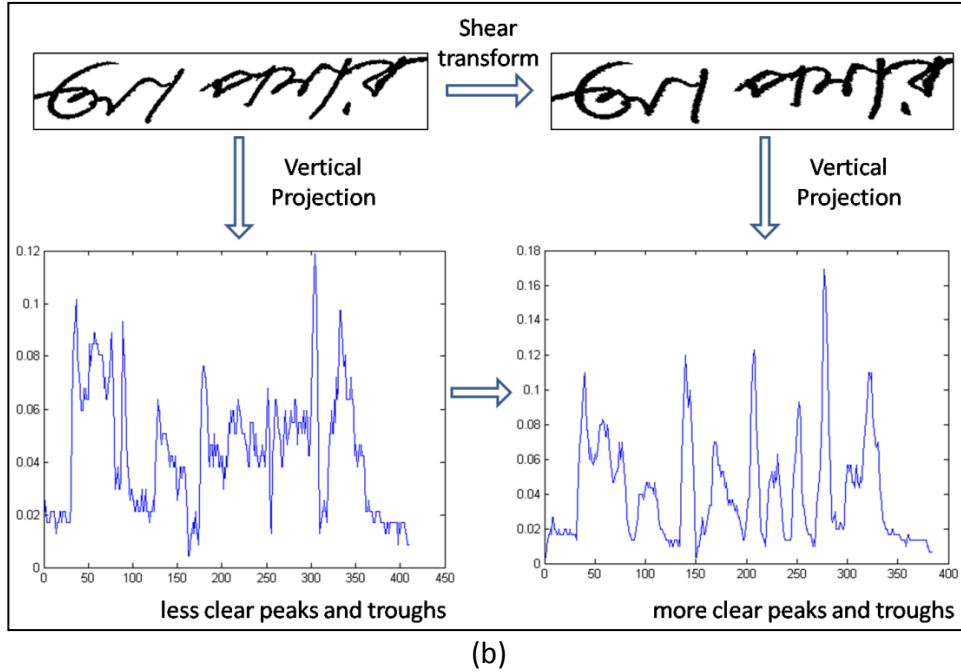

(b)

**Fig.6:(a) Skew correction using water reservoir. Water reservoirs are computed and checked for consideration according to reservoir height. Next a regression analysis is performed on depth-points of the reservoirs and skew angle is obtained. (b) Slant correction was performed using vertical projection analysis.**

## 4. Proposed Zone-based Word Recognition Approach

In this section we present our zone segmentation based Indic word recognition framework. We have used the combination of HMM and SVM based classification for handwritten word recognition. After performing the preprocessing tasks, the word image is passed through zone segmentation module. Unlike, traditional Indic handwritten word recognition approaches, by segmenting the words into 3 zones we reduce the number of basic units for character recognition (discussed in Section 2). After segmenting into zones, recognition of middle zone components is performed using HMM. Upper and lower zone components are recognized using SVM classifier and finally zone wise results are combined to get final results. The proposed zone-based recognition system outperforms traditional character-wise word recognition approach. We also demonstrate that the refinement in zone-segmentation improves the recognition performance. The architecture of the system has been shown in Fig.7 with a Bangla word example. The details of these steps are discussed briefly in following subsections.

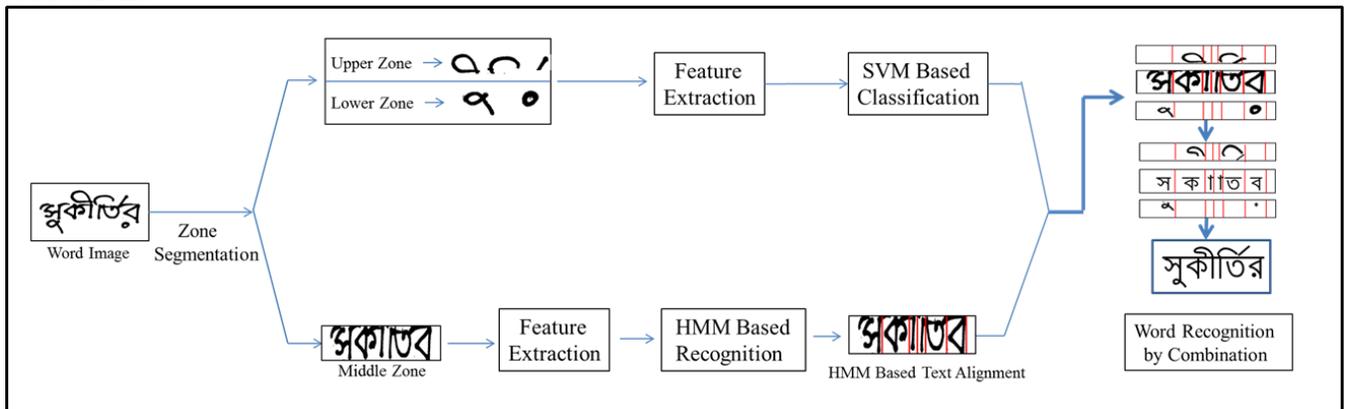

**Fig.7: Outline of our word recognition framework**

## 4.1. Zone Segmentation

After rectifying the skew and slant defects, the words are segmented into 3 zones: upper, middle and lower zones. For this purpose, the *Matra* in a word is first detected. Due to complex writing, exact *Matra* detection is not easy always. Hence we detect possible regions of *Matra*. These are explained as follows.

**4.1.1. Matra Row Detection:** In literature of printed word recognition [10], *Matra* is usually determined by projection analysis in horizontal direction and considering the row with highest peak. But, due to the free flow nature of handwriting the *Matra* is rarely a perfect straight line. It is often curvy and broken. To determine the estimated location of *Matra*, we considered three different row information for locating the approximated row in the word image. Next, the best one among these three is chosen. First row, denoted as $R_1$, is the highest peak determined by projection analysis of the word in horizontal direction. Second row, $R_2$, is the row calculated from depth-points of water reservoirs where the sum of the squares of the distance between this row and each of the depth-points is minimum.
The third row, $R_3$ is computed as follows. Since, the upper zone of Devanagari/Bangla script contain fewer components than that in the middle zone, the portions below the *Matra* will be more dense than that of above the *Matra*. Hence, there will be a sharp decline in projection peak in upper half of the word while moving from below the *Matra* to above the *Matra*. We mark the row where a sharp decline in projection is observed as the third estimated location ($R_3$) of *Matra*. Finally, the *Matra* row has been detected by following rules. These rows are shown in Fig. 8.

$$\textbf{Approximate headline row} = \begin{cases} R_1, & \text{if } |R_1 - R_2| \leq T_h \text{ or } |R_1 - R_3| \leq T_h \\ R_2, & \text{if } |R_2 - R_3| \leq T_h \text{ and } |R_1 - R_2| > T_h \text{ and } |R_1 - R_3| > T_h \\ R_3, & \text{otherwise} \end{cases} \quad ...(1)$$

Where, $T_h = H/10$, is the threshold and $H$ is the height of the word image; $H$ is calculated by taking mode of height list taken from top and bottom most pixels of each column of the word image. We have noticed that, the location of *Matra* row has been efficiently detected with this rule in most of the word images.

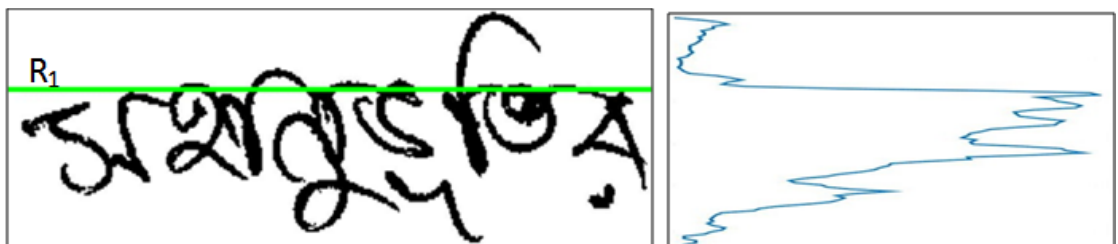

(a)

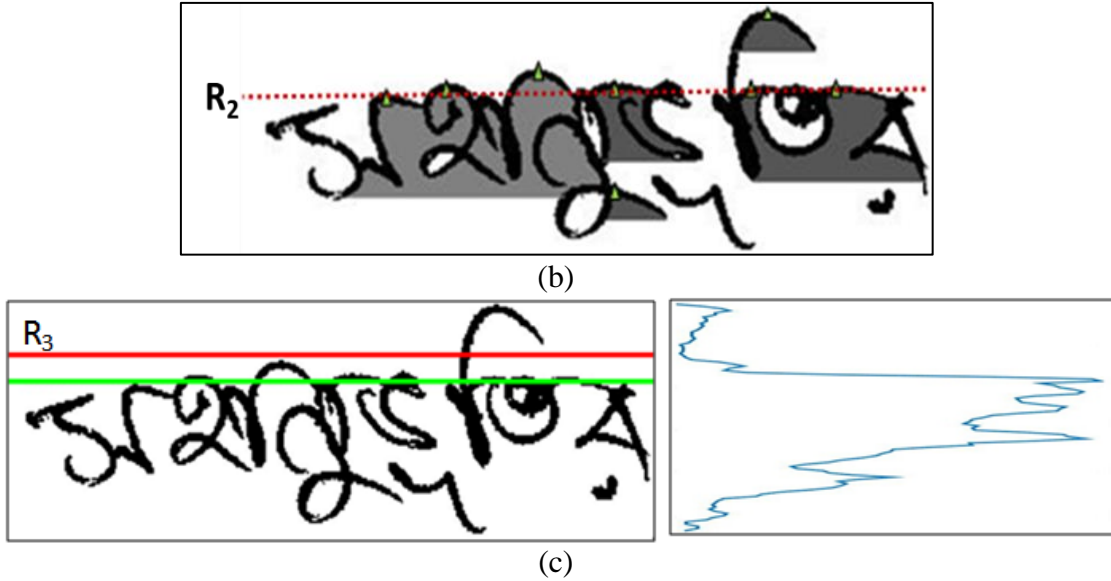

(b)

(c)

**Fig.8:** An example of Bangla word and corresponding rows for *Matra* detection. (a) Row $R_1$ is marked along with the projection analysis of the word in horizontal direction, (b) Bottom water reservoirs (shaded in gray) are shown. Depth points are denoted in green peaks. Row $R_2$, obtained by regression analysis of these peaks is marked in red dotted line. (c) Row $R_3$ is marked by black line.

**4.1.2. Upper Zone Segmentation:** After estimating the *Matra* row, we create a window of *Matra* region ($W_M$) of height $4 \times S_w$ keeping the *Matra* in middle, where $S_w$ is the stroke width (discussed in Section 3). It is noted that the curvilinear *Matra* resides in $W_M$ in more than 98% of the words from experiment dataset. Next, we extract the skeleton of the word image and find the high curvature points, junction points, and end points of the skeleton image (See Fig. 9). These points are marked as 'P'. Now, we find the lines between consecutive 'P's in horizontal direction within $W_M$. If any line-segment emerging from point 'P' and crosses $W_M$, we consider it as a character-portion and hence discard it. Only those line-segments between 'P's which are passed within $W_M$ are considered. If more than one pixel is found in a single column we consider the upper most pixel. In some words *Matra* may be broken and discontinuous. There, we join the two nearest *Matra* pixels using standard Bresenham algorithm. Next the modifiers in the upper zone are marked by checking the upper portions of *Matra*.

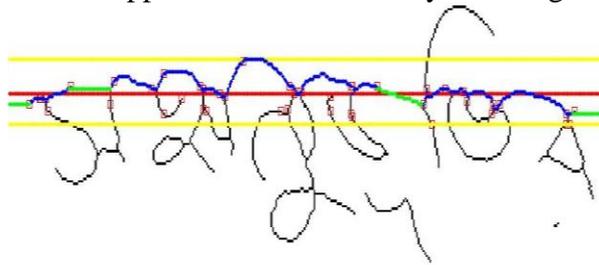

**Fig.9:** Examples of Matra and Matra region detection. (a) Red line is the Matra row. Yellow lines specify the *Matra* region. Red squares represent high curvature points, corners and junction points within this frame. Blue line denotes the detected Matra. Green line joins the nearest Matra pixels in case of broken Matra. (For better visibility please see the soft copy of the PDF version of the paper)

**4.1.3. Lower Zone Segmentation:** In our earlier approach [19], to detect the modifiers in lower zone, we marked the baseline which separated the middle zone from lower zone by observing a sharp decline in the busy zone in lower half of the image. This approach may fail sometimes in situation, when the baseline that separates the middle zone from the lower zone is difficult to locate. If the letters of the word are irregular in size, or there exist many modifiers in lower zones, then we may not find any sharp decline in projection peak between the middle zone and lower zone of the image. To have an idea we show an example in Fig. 10 where due to complex writing style, the lower zone detection becomes difficult using projection analysis.

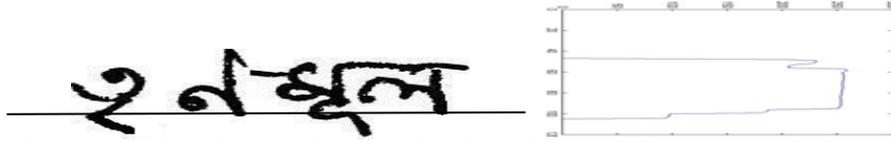

**Fig.10: Examples of cases where baseline detection may fail to segment the lower zone**

To overcome this problem we include a shape matching based algorithm for modifier extraction in this paper. To segment the lower zone modifiers, we search for modifiers in lower half of the image by shape matching. To do so, we find the touching location of modifier by skeleton analysis and separate them from middle zone. If the residue shape components is matched with any of the lower zone characters with high matching confidence, that part is separated from middle zone. The flowchart of lower zone modifier separation is described in Fig.11. The segmentation is discussed below in details.

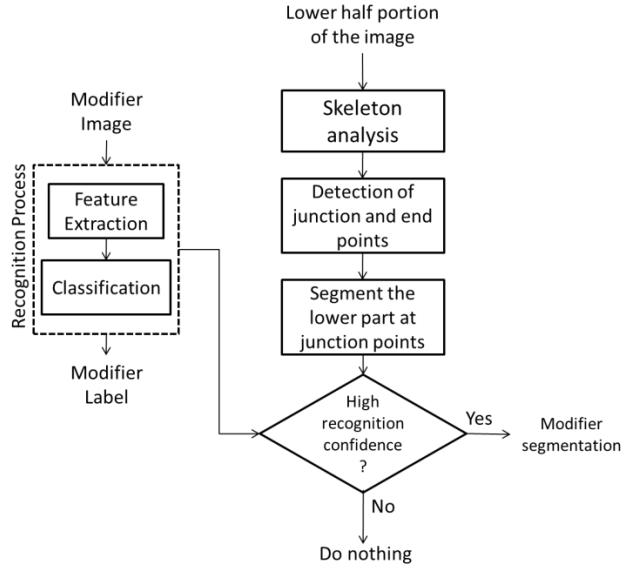

**Fig.11: Modifier segmentation from lower portion of the image**

Let $M_1$ be the word image. The skeleton image of $M_1$ is first obtained and the junction and end points in skeleton image are detected. Let $L_1$ be the lower half of the image $M_1$. Using connected component (CC) analysis in $L_1$, the components which are not connected to $M_1$ are detected as modifiers. 8-connection connectivity was used in CC analysis. Some of the lower modifiers can be touching to $M_1$. To separate these modifiers, the skeleton of the components are traced from lower end points. The intuition behind tracing is that usually, all lower modifiers have one end point in lower half of the word, and except the modifier ॐ which does not have junction point. In case of ॐ, the junction point is always a part of a loop. If loop is found, we continue the tracing to detect the next junction point for segmentation. If more than one endpoint is found in a column the lowermost of it is considered in segmentation analysis. After segmentation, given an image portion, we compute the recognition confidence using SVM classifier to obtain the corresponding class label. Details of feature extraction and recognition using SVM are given in Section 4.3.2. The probability score is calibrated using Platt scaling from SVM score [23, 24]. Radial Basis Function (*RBF*) kernel was chosen in SVM for better performance.If a component is recognized by the SVM with a high confidence (more than 0.6), we consider it as a modifier. If the difference between the top two recognition scores of a component is high, it is also considered as a modifier. In Fig. 12, we show an example of lower zone separation from a Bangla word image.

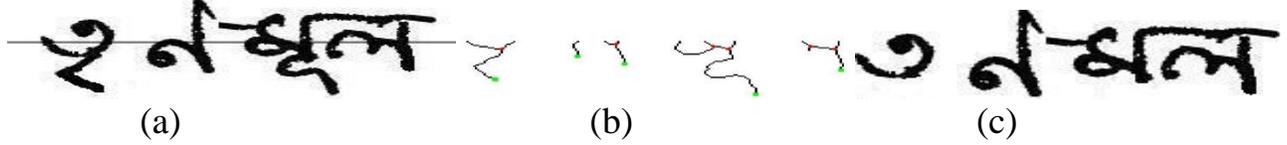

(a)            (b)            (c)

**Fig.12: (a) $M_1$ with horizontal line separating the lower half from upper half. (b) Skeleton of $L_1$ with end points marked in green and junction points marked in red. (c) Lower zone separated out from word.**

## 4.2. Feature Extraction

The middle zone is the primary portion in Devanagari and Bangla word region where characters are often touching with each other. We apply Hidden Markov Model (HMM) based stochastic sequential classifier for recognizing the touching components in this zone.

For HMM-based middle zone recognition, we have developed an efficient feature extraction technique PHOG using multi-resolution HOG features [25]. To measure its effectiveness we have implemented 4 different state-of-the-art approaches and compared their performances. In our previous work[19] we used the LGH (Local Gradient Histogram) feature for feature extraction. Here, we study other features, mainly, profile feature, GABOR feature and G-PHOG feature (combination of Gabor and PHOG) for middle zone recognition. These features are briefly described below.

### 4.2.1. PHOG Feature:
PHOG[26] is the spatial shape descriptor which gives the feature of the image by spatial layout and local shape, comprising of gradient orientation at each pyramid resolution level. To extract the feature from each sliding window, we have divided it into cells at several pyramid level. The grid has $4^N$ individual cells at *N* resolution level (i.e. *N*=0, 1,2..).Histogram of gradient orientation of each pixel is calculated from these individual cells and is quantized into *L* bins. Each bin indicates a particular octant in the angular radian space.

The concatenation of all feature vectors at each pyramid resolution level provides the final PHOG descriptor. L-vector at level zero represents the L-bins of the histogram at that level. At any individual level, it has $L \times 4^N$ dimensional feature vector where N is the pyramid resolution level (i.e. N=0, 1, 2….). So, the final PHOG descriptor consists of $L \times \sum_{N=0}^{N=K} 4^N$ dimensional feature vector, where *K* is the limiting pyramid level. In our implementation, we have limited the level (N) to 2 and we considered 8 bins (360º/45º) of angular information. So we obtained (1×8) + (4×8) + (16×8) = (8+32+128) = 168 dimensional feature vector for individual sliding window position (See Fig.13).

### 4.2.2. LGH Feature:
LGH feature [27], proposed by Rodriguez and Perronin, was similar to HOG feature [25] for object recognition. A sliding window of fixed width is being shifted from left to right of the word image with an overlapping between two consecutive frames. Next, feature is computed from each sliding window by dividing into 4x4 cells. From each cell, Histogram of Gradient(with 8 bins) is computed and the final feature vector is the concatenation of 16 histograms which gives a 128 dimensional feature vector for each sliding window position. The image is smoothed by Gaussian filter before feature extraction for better gradient information.

### 4.2.3. GABOR Feature:
The GABOR features has been applied successfully in character and word recognition [28]. Here for our work Gabor filtering in four orientations (0°, 45°, 90° and 135°) is applied and then we used the magnitude as the response for feature extraction. After filtering, the image frame is divided equally into 12 rows. Next, we concatenate the features in each grid to have 48dimensional Gabor features.

### 4.2.4. G-PHOG Feature:
We have also made an experiment with a combination of Gabor and PHOG features called G-PHOG feature. The idea of G-PHOG is motivated from the work of [29] where Gabor feature has been combined to improve the result. From the experiment using G-PHOG feature, it is noted that the efficiency of Gabor feature can be improved by combining it with PHOG descriptor.

### 4.2.5. Marti-Bunke Feature:
The profile feature proposed by [30]used extensively for Latin script recognition, consists of nine features computed from foreground pixels in each image column. Three global features are used to capture the fraction of foreground pixels, the centre of gravity and the second order moment. Remaining six local features comprise of the position of the upper and lower profile, the number of foreground to background pixel transitions, the fraction of foreground pixels between the upper and lower profiles and the gradient of the upper and lower profile with respect to the previous column, which provides dynamic information.

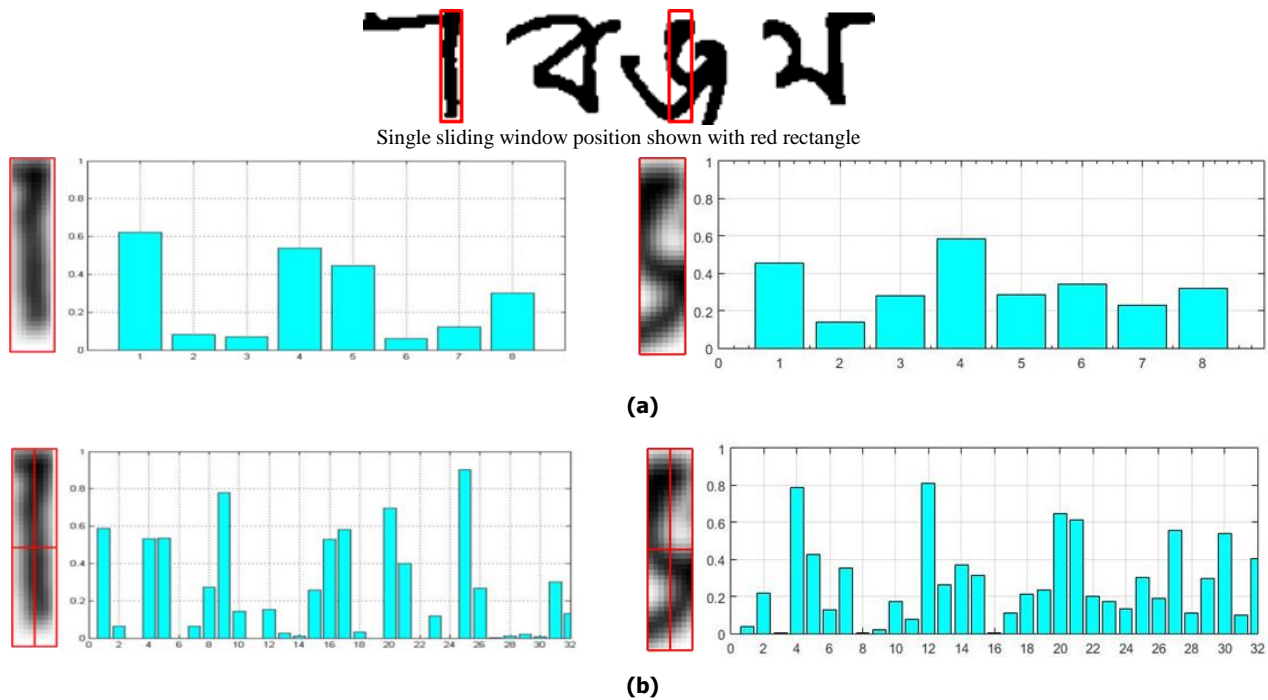

Single sliding window position shown with red rectangle

(a)

(b)

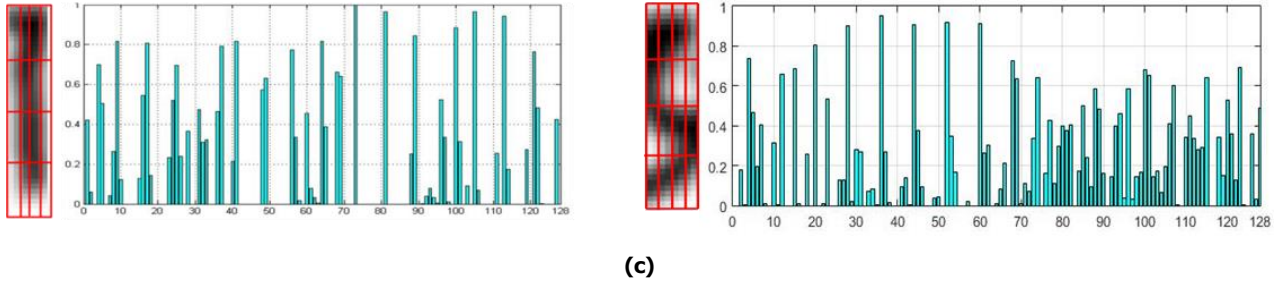

**(c)**

**Fig.13: PHOG feature extraction from a Bangla word image where feature vector at each resolution level is concatenated to give the final PHOG descriptor. Gaussian smoothing applied and feature vector at (a) 0th pyramid resolution level (b) 1st pyramid resolution level and (c) 2nd pyramid resolution level**

## 4.3. Recognition

### 4.3.1. Middle Zone Recognition using Hidden Markov Model

We extracted each of the above feature descriptors using sliding window and apply HMM for word recognition. The feature vector sequence is processed using left-to-right continuous density HMMs [30]. One of the important features of HMM is the capability to model sequential dependencies. An HMM can be defined by initial state probabilities $\pi$, state transition matrix A $=[a_{ij}]$, i, j=1,2,…,N, where $a_{ij}$ denotes the transition probability from state i to state j and output probability $b_j(O_K)$ modeled with continuous output probability density function . The density function is written as $b_j(x)$, where x represents *k* dimensional feature vector. A separate Gaussian mixture model (GMM) is defined for each state of model. Formally, the output probability density of state *j* is defined as

$$b_j(x) = \sum_{k=1}^{M_j} c_{jk}\, \mathcal{N}(x, \mu_{jk}, \Sigma_{jk}) \qquad (1)$$

where, $M_j$ is the number of Gaussians assigned to *j*. and $\mathcal{N}(x,\mu,\Sigma)$ denotes a Gaussian with mean $\mu$ and covariance matrix $\Sigma$ and $c_{jk}$ is the weight coefficient of the Gaussian component k of state *j*. For a model $\lambda$, if O is an observation sequence $O=(O_1, O_2,..,O_T)$ which is assumed to have been generated by a state sequence $Q=(Q_1, Q_2,..,Q_T)$, of length *T*, we calculate the observations probability or likelihood as follows:

$$P(O,Q|\lambda) = \sum_Q \pi_{q1} b_{q1}(O_1) \prod_T a_{q_{T-1}\, q_T} b_{q_T}(O_T) \qquad (2)$$

Where $\pi_{q1}$ is initial probability of state 1. In the training phase, the transcriptions of the middle zone of the word images together with the feature vector sequences are used in order to train the character models. The recognition is performed using the Viterbi algorithm. For the HMM implementation, we used the HTK toolkit [31]. The parameters like, numbers of Gaussian Mixture and state are fixed according to validation data.

### 4.3.2 Support Vector Machine (SVM) and Modifier Recognition

The isolated components which were included in upper and lower zones are segmented using connected component (CC) analysis and next they are recognized and labelled as text characters. If the components are broken, an algorithm due to Roy et al. [32] is applied to join the broken contours in the image. Upper zone modifiers like ⌒ , ⌒ , ⌒ , ∕ , ‿ (for Bangla) and ⌒, ⌒, ⟍, ⟍, ♦, ∪ (for Devanagari) and Lower zone modifiers like ় ,ু ,ৃ,ূ ,ৢ , ● (for Bangla) and ु, ृ,ॄ (for Devanagari) are separately considered for classification so that the chances of error can be minimized. After resizing the images to 150x150, PHOG feature of vector length 168 is extracted from upper and lower zone modifiers. PHOG feature is considered as it provided better result in the experiment. Next, Support Vector Machine (SVM) classifier [33] has been used to classify these components.

SVM classifier has been chosen here as it has successfully been applied in a wide variety of classification problems[33]. Given a training database of M data: $\{x_m|\ m=1,..,M\}$, the linear SVM classifier is defined as:

$$f(x) = \sum_j \alpha_j x_j + b$$

Where, $x_j$ is the set of support vectors and the parameters $\alpha_j$ and $b$ have been determined by solving a quadratic problem. A linear kernel can be used to classify data which have fewer variations. But changing the kernel function to Radial Basis Function (*RBF*)was a better choice in our experiment study to classify upper and lower zone modifiers for recognition.LIBSVM toolbox [23] was used for SVM based learning. We used grid search technique to optimize the gamma and multiplier parameter in the library. In grid-search we started with a coarse value of parameters, and then a finer grid around the best parameter values was used.The SVM prediction result is used afterwards to merge with middle zone recognition results to form the entire word.

### 4.3.3. Combination of Zone-wise Recognition Results

In this section, the details of the modifier alignment and combination with the middle zone results are discussed. For estimating the boundaries of the characters in the middle zone of a word, Viterbi Forced Alignment *(FA)* has been used in the middle-zone of the word. With the embedded training of FA, the optimal boundaries of the characters of the middle-zone are found. After obtaining the character boundaries in the middle zone, the respective boundaries are extended in the upper and lower zones to associate characters present in upper and lower zones with the middle zones characters. This is one hypothesis for characters segmentation and combination of a given word. Similarly, we generate N such hypothesis using N-best Viterbi list obtained from middle zone of the word.

The score to generate a hypothesis is calculated based on the recognition results of middle zone. For a given word image *(X),* its score is calculated based on a lexicon *(W)* of the middle zone characters and it is theposterior *P(W|X)*. Using logarithm in Bayes' rule we get

$$\log p(W|X) = \log p(X|W) + \log p(W) - \log p(X)$$

From these scores N-best hypothesis are chosen. Now among these N-best choices, the best hypothesis is chosen combining upper and lower zone information discussed as follow.

After computing the zone-wise recognition results (upper and lower zone modifiers are recognized by SVM and middle zone characters by HMM of a word (X) and recognized character labels are obtained) the labels of upper and lower zones are associated with labels of middle zone. The association of character labels can be considered as a path-search problem to find the best matching word where each character label will be used only once. In our framework, the association is performed as follows. Let, the recognition labels of middle zone characters be $C_{M\_1}$, $C_{M\_2}$ ... $C_{M\_N}$ where $N$ is the number of characters obtained in middle zone. Also, let the recognition labels of upper zone characters and lower zone characters be $C_{U\_1}$, $C_{U\_2}$,...,$C_{U\_N}$ and $C_{L\_1}$, $C_{L\_2}$,...,$C_{L\_N}$ respectively (please note that in most of the cases, number of upper and lower zone characters will be less than middle zone characters). Let these zone-wise character results are obtained and stored in 3 arrays $C_U[]$, $C_L[]$, and $C_M[]$ respectively (See Fig.14(b)). A middle zone character ($C_{M\_i}$) generally be associated to its corresponding upper ($C_{U\_i}$) and lower zone ($C_{L\_i}$) modifies. After association, ideally the whole word will be $W_T= C_1, C_2...C_N$ where $C_i= F\_i(C_{M\_i}, C_{U\_i}, C_{L\_i}), i = 1, 2, 3...N$ and $F\_i$ is an association function of middle zone character ($C_{M\_i}$) with $C_{U\_i}$ and $C_{L\_i}$. But, due to complex handwriting styles, some upper/lower zone modifiers may not appear exactly above and below of their middle zone character (see Fig. 14(a)). To handle such situation, a more flexible association rule is proposed here. In this modified association rule, a middle zone character ($C_{M\_i}$) is associated not only its exactly upper ($C_{U\_i}$) and lower ($C_{L\_i}$) zone modifiers but also $C_{M\_i}$ associates with one modifier ($C_{U\_i-1}$), ($C_{U\_i}$) and ($C_{U\_i+1}$) from upper zone and one modifier ($C_{L\_i-1}$), ($C_{L\_i}$) and ($C_{L\_i+1}$) from lower zone. Thus our modified association rule of middle zone character ($C_{M\_i}$) becomes $F\_i(C_{M\_i},C_{U\_i},C_{U\_i+1},C_{U\_i-1},C_{L\_i},C_{L\_i+1},C_{L\_i-1})$. Similarly for each middle zone character we find associated upper and lower zone characters and hence associated words are obtained. For each word we may have several associated words $W_T^1$, $W_T^2$,..., $W_T^S$, where $s$ is total number of words formed. Each associated word ($W_T^j$) is matched with the lexicon ($L$) and best matched associated word is the combined zone-wise result of the word ($X$). The similarity score in lexicon matching is obtained using Levenshtein distance [34]. By Levenshtein distance matching, we find errors (characters that are not same) between two sequence of characters. The errors are considered as the difference from substitution, insertions, and deletions operation. This string matching algorithm is solved using dynamic programming (DP). Thus, we obtain a distance score for each associated word along with its word selected from lexicon. The scores are next sorted and the lexicon word with minimum score is considered as best result. In Fig.14(b), we illustrate this association process to get combined result. Algorithm 1 details the steps of the combination of zone wise results to find the best matching result.

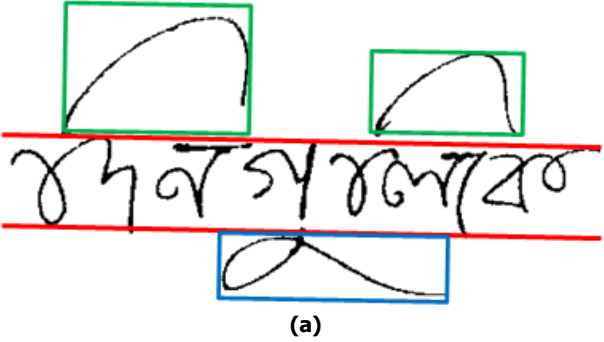

(a)

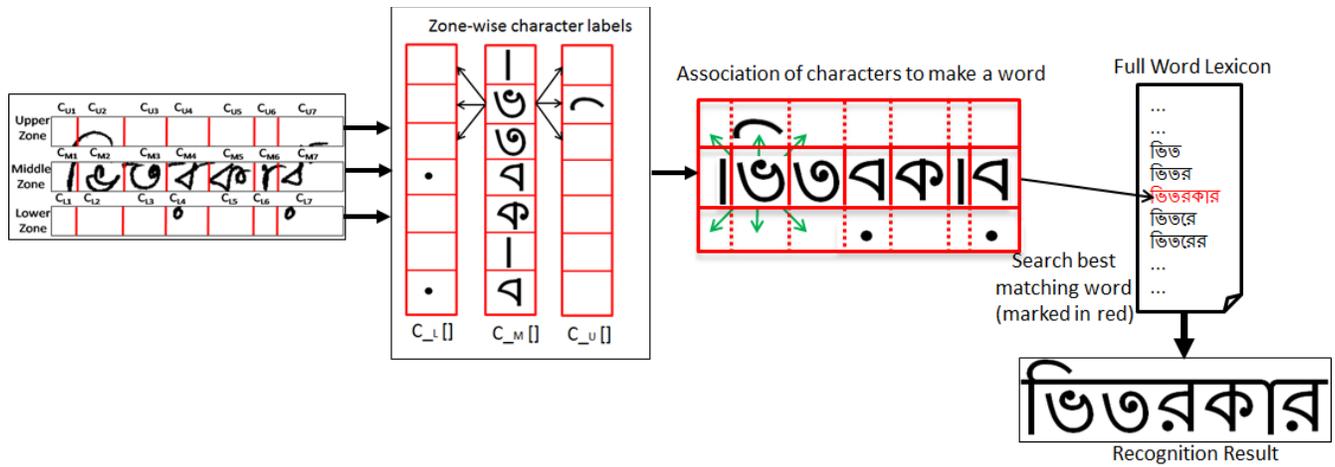

**(b)**

**Fig.14:** (a) Upper (Green color) and lower (blue color) modifiers are marked in a Bangla word image. Middle zone characters are marked within red box. Note that, these upper and lower modifiers cover more than one character in middle zone. (b) Example of character segmentation in middle zone result applied to upper and lower zones for modifier separation purpose. After obtaining zone-wise character results $C_M[]$, $C_L[]$, and $C_U[]$, characters in 3 zones are associated to form a word. Arrow in green colour denotes the possible association of a middle zone character ($2^{nd}$ character in middle zone) with upper and lower zone characters. Next, the associated word is matched with lexicon list to find the best matching result.

---

**Algorithm 1.** Combination of Zone wise results to form a full word

**Require:** Recognition of zone wise results. Middle zone results ($C_{M\_1}$, $C_{M\_2}$ ... $C_{M\_N}$) using HMM and upper ($C_{U\_1}$, $C_{U\_2}$ ... $C_{U\_N}$) and lower ($C_{L\_1}$, $C_{L\_2}$ ... $C_{L\_N}$) zone results using SVM.
**Ensure:** Full word recognition result

**Step 1:** Estimate boundaries of middle zone characters ($C_{M\_i}$) using Viterbi Forced Alignment.

**Step 2:** Extension of the corresponding character boundary of middle zone in upper and lower zone.

**Step 3:** Associate each middle zone character with upper and lower zone character to form a word $W_T^j$. The association rule is that, a middle zone character $C_{M\_i}$ can associate with any character in upper zone from ($C_{U\_i}, C_{U\_i+1}, C_{U\_i-1}$) and any character in lower zone from ($C_{L\_i}, C_{L\_i+1}, C_{L\_i-1}$) respectively. Note that, each character will be used only once in this association

**Step 4:** All associated words ($W_T^j$) are matched with lexicon (*L*) using Levenshtein distance algorithm and a similarity score is obtained corresponding to each word $W_T^j$.

**Step 5:** The lexicon string with minimum distance is considered as the best matching and it is taken as the final combined recognition result.

---

Due to complex handwriting variation in Indic script, the Viterbi forced alignment algorithm may not always give proper character segmentation in middle zone. To improve the middle zone character segmentation, "Water reservoir" [21] is used. To do so, first, the character segmentation points obtained by Viterbi algorithm are noted. Next, the bottom reservoirs (as explained in Section 3.1) are computed in that word image. The reservoirs having low height are discarded and the deepest points of the rest of the reservoirs are found. The character segmentation using Viterbi is next shifted towards the nearest water reservoir deepest point. For illustration, we show in Fig. 15 the improvement of character segmentation using water reservoir concept. Note that the segmentation lines in blue color in Fig.15(b) were not correct using Viterbi algorithm. After using water reservoir, these segmentation errors are rectified (shown in Fig.15(d)).

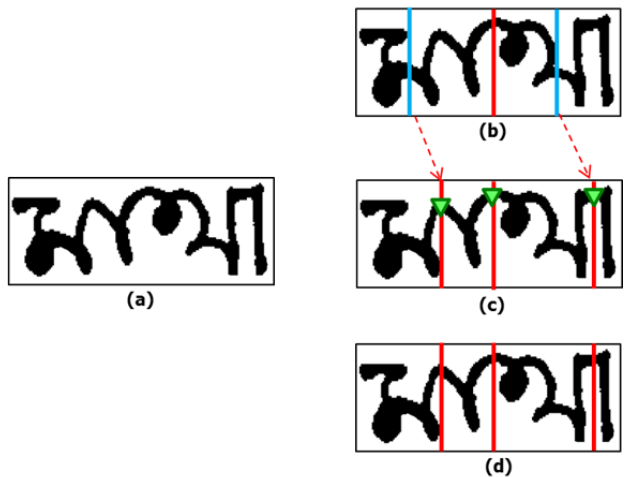

**Fig. 15.** Improvement of character segmentation using Water reservoir approach. (a) Middle zone word image (b) Viterbi algorithm based segmentation lines are shown. Here, correct segmentation line is indicated by red and improper segmentation is indicated by blue line (c) Deepest points of the bottom water reservoir are indicated by green triangle and segmentation lines through these points are marked. (d) Modified character segmentation lines are marked. The incorrect lines from Viterbi algorithm are moved to their corresponding nearest deepest points obtained in (b).

## 5. Experimental Results and its Analysis

For experiment of the handwritten word recognition scheme we collected two sets of data from Bangla and Devanagari scripts. For Bangla scripts we collected a total of 17,091 handwritten word samples. These words were considered from 60 handwritten document images from individual of different professions. Among these word images 11,253 images are used for training, and 1,982 word images as validation data and rest 3,856 samples are for testing. A list of 1,547 Bangla words is considered in the lexicon. For Devanagari script, we have collected a total of 16,128 handwritten word images, out of which 10,667 word images are used for training, 1,872 word images for validation and rest 3,589 images are used for testing. Details of the data are shown in Table I. Datasets of both Bangla and Devanagari scripts are made available online for further research [36]. We have considered 1,957 Devanagari words in the lexicon. The distribution of word according to character length is shown in Fig.16. It was noted that words having length of 4 characters were largest in both Bangla and Devanagari dataset.

**Table I: Description of data details with number of word images for experiment evaluation**

|            | Training Data | Validation Data | Testing Data | Total  |
|------------|---------------|-----------------|--------------|--------|
| Bangla     | 11,253        | 1,982           | 3,856        | 17,091 |
| Devanagari | 10,667        | 1,872           | 3,589        | 16,128 |

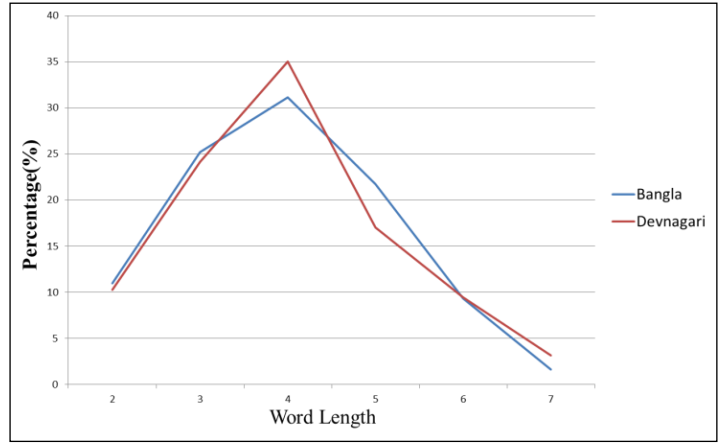
**Fig.16: Percentage of words according to their length**

**5.1. Middle zone recognition**

After applying the pre-processing in each word images, horizontal zone segmentation was performed to segment the word into 3 different zones. The middle zone is next processed using HMM-based recognition. Due to zone segmentation, the size of lexicon in middle zone recognition is reduced. We obtained 1,518 and 1,894 words in Bangla and Devanagari, respectively. We considered continuous density HMMs with diagonal covariance matrices of GMMs in each state. Five different features are extracted from the training samples. A number of Gaussian mixtures and state numbers were tested on validation data in both Bangla and Devanagari dataset. We noted that with zone segmentation the number of character-component class in HMM for Bangla (Devanagari) is reduced from 124 to 42 (116 to 40) in our dataset.

The stacked column charts in Fig.17 and Fig.18 show the performance using different Gaussian Mixture numbers and top N choices in validation data. Token passing algorithm [31] was used in our algorithm for computing n-best choices. It is observed that the 32 Gaussian Mixture PHOG feature provides the best results achieving up to 92.98% (Bangla) and 94.59% (Devanagari) accuracy with top 5 choices in validation data. PHOG outperformed the other features in middle zone recognition. The nearest LGH feature achieved 91.21% (Bangla) and 92.45% (Devanagari) accuracy. Both PHOG and LGH result were obtained with sliding window size of 40x6 and step-size of 3. The performance with varying state number is shown in Fig.19 and it is noted that with state number 8 the best result is obtained. Since the middle zone recognition results are combined with upper and lower zone modifiers to get the final word level, we have analyzed upto top 5 choice results for re-ranking purpose and considered all of them with combination of upper and lowers zone modifiers. Performance accuracies for different Gaussian and state numbers are shown in tabular format in Table II & Table III.

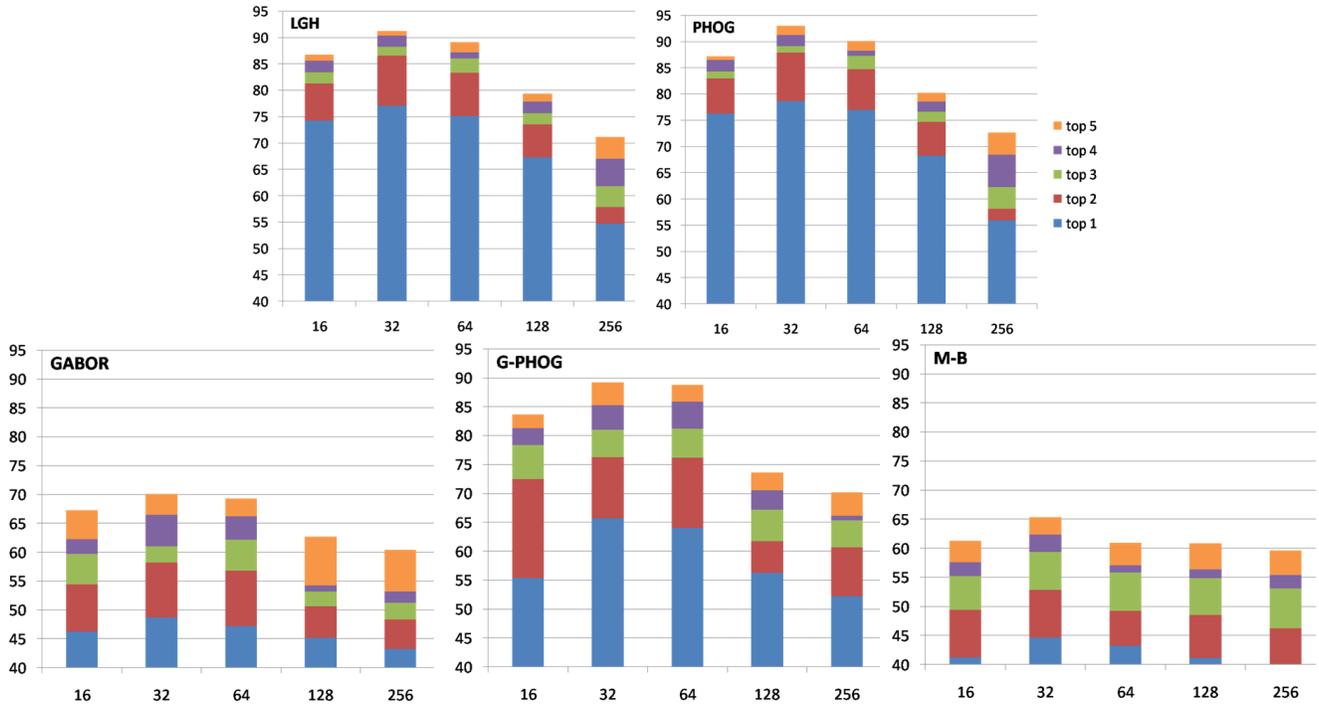

**Fig.17:** Performance accuracy plotted against Gaussian Mixtures with Top N choices as parameter for LGH, PHOG, GABOR, G-PHOG, Marti-Bunke (M-B) feature for Bangla handwriting recognition.

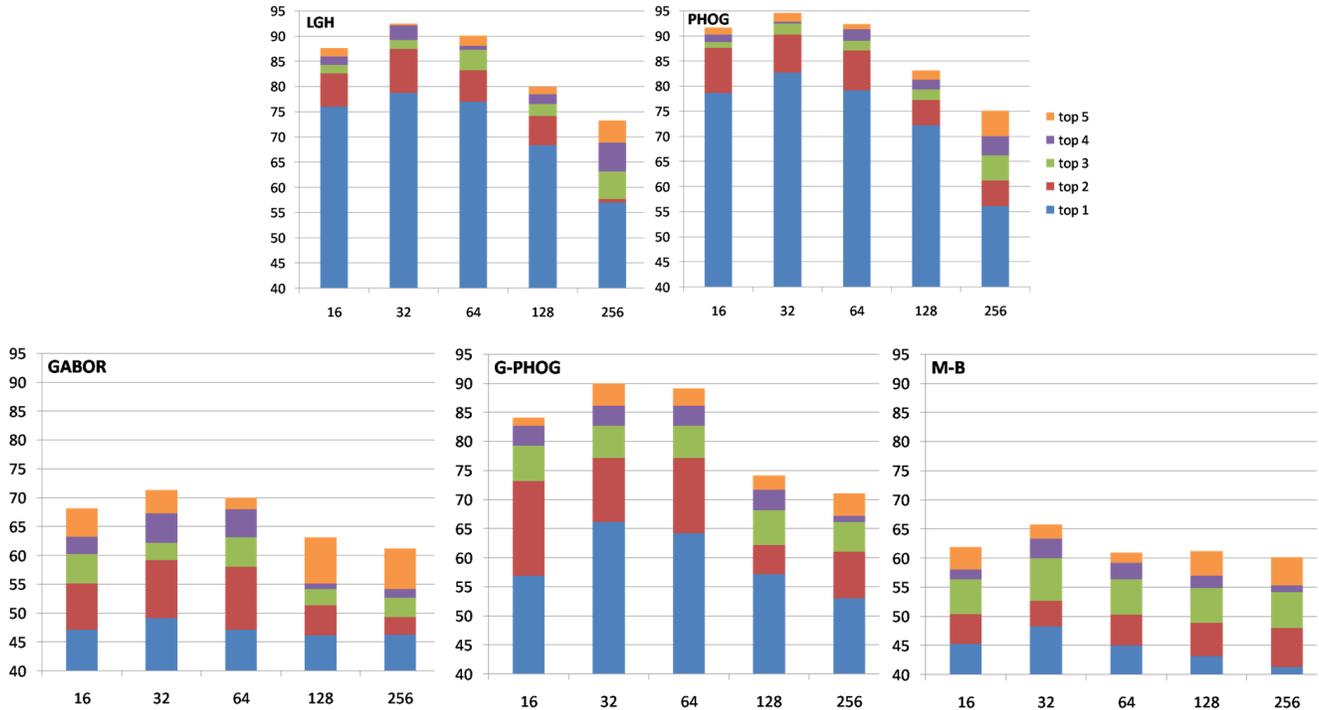

**Fig.18:** Performance accuracy plotted against Gaussian Mixtures with Top N choices as parameter for LGH, PHOG, GABOR, G-PHOG, Marti-Bunke(M-B) feature for Devanagari handwriting recognition.

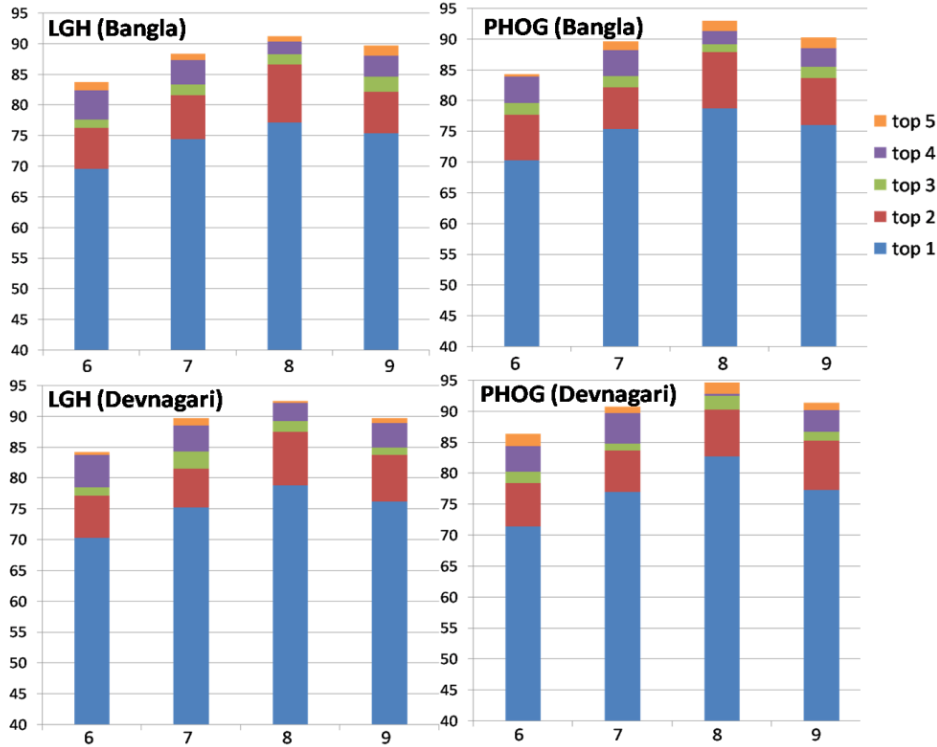

**Fig.19: Comparison between different state numbers with LGH&PHOG features**

**Table II: Performance accuracy in tabular format for different Gaussian numbers**

| Feature | LGH | | | | | PHOG | | | | | GABOR | | | | |
|---|---|---|---|---|---|---|---|---|---|---|---|---|---|---|---|
| **Bangla** | | | | | | | | | | | | | | | |
| Gaussian# | top 1 | top 2 | top 3 | top 4 | top 5 | top 1 | top 2 | top 3 | top 4 | top 5 | top 1 | top 2 | top 3 | top 4 | top 5 |
| 16 | 74.26 | 81.29 | 83.42 | 85.58 | 86.76 | 76.31 | 82.99 | 84.26 | 86.48 | 87.15 | 46.22 | 54.45 | 59.68 | 62.24 | 67.26 |
| 32 | 77.09 | 86.61 | 88.25 | 90.35 | 91.21 | 78.69 | 87.89 | 89.15 | 91.28 | 92.98 | 48.69 | 58.23 | 61.03 | 66.54 | 70.12 |
| 64 | 75.11 | 83.31 | 86.05 | 87.22 | 89.14 | 77.01 | 84.69 | 87.26 | 88.21 | 90.08 | 47.23 | 56.78 | 62.14 | 66.25 | 69.31 |
| 128 | 67.29 | 73.58 | 75.66 | 77.85 | 79.34 | 68.25 | 74.69 | 76.58 | 78.56 | 80.25 | 45.15 | 50.61 | 53.22 | 54.29 | 62.67 |
| 256 | 54.67 | 57.91 | 61.84 | 67.01 | 71.14 | 55.96 | 58.15 | 62.25 | 68.45 | 72.69 | 43.25 | 48.32 | 51.31 | 53.19 | 60.39 |
| **Devanagari** | | | | | | | | | | | | | | | |
| Gaussian # | top 1 | top 2 | top 3 | top 4 | top 5 | top 1 | top 2 | top 3 | top 4 | top 5 | top 1 | top 2 | top 3 | top 4 | top 5 |
| 16 | 75.99 | 82.65 | 84.31 | 85.94 | 87.61 | 78.69 | 87.59 | 88.79 | 90.25 | 91.69 | 47.16 | 55.15 | 60.29 | 63.23 | 68.18 |
| 32 | 78.77 | 87.49 | 89.17 | 92.15 | 92.45 | 82.69 | 90.25 | 92.45 | 92.78 | 94.59 | 49.15 | 59.18 | 62.18 | 67.26 | 71.36 |
| 64 | 77.01 | 83.26 | 87.29 | 88.07 | 90.08 | 79.2 | 87.12 | 89.01 | 91.36 | 92.36 | 47.16 | 58.02 | 63.18 | 67.97 | 70.02 |
| 128 | 68.33 | 74.18 | 76.54 | 78.5 | 79.99 | 72.26 | 77.23 | 79.31 | 81.3 | 83.12 | 46.19 | 51.39 | 54.17 | 55.18 | 63.18 |
| 256 | 56.98 | 57.69 | 63.12 | 68.84 | 73.25 | 56.12 | 61.23 | 66.19 | 70.02 | 75.15 | 46.25 | 49.29 | 52.69 | 54.14 | 61.19 |

| Feature | G-PHOG | | | | | M-B | | | | |
|---|---|---|---|---|---|---|---|---|---|---|
| **Bangla** | | | | | | | | | | |
| Gaussian# | top 1 | top 2 | top 3 | top 4 | top 5 | top 1 | top 2 | top 3 | top 4 | top 5 |
| 16 | 55.39 | 72.49 | 78.34 | 81.29 | 83.67 | 41.19 | 49.39 | 55.24 | 57.59 | 61.31 |
| 32 | 65.69 | 76.23 | 81.01 | 85.25 | 89.22 | 44.62 | 52.87 | 59.36 | 62.37 | 65.36 |
| 64 | 63.99 | 76.21 | 81.17 | 85.9 | 88.77 | 43.28 | 49.27 | 55.81 | 57.07 | 60.91 |
| 128 | 56.26 | 61.78 | 67.21 | 70.58 | 73.66 | 41.14 | 48.54 | 54.87 | 56.36 | 60.87 |
| 256 | 52.22 | 60.69 | 65.39 | 66.12 | 70.21 | 39.31 | 46.26 | 53.14 | 55.39 | 59.64 |
| **Devanagari** | | | | | | | | | | |
| Gaussian # | top 1 | top 2 | top 3 | top 4 | top 5 | top 1 | top 2 | top 3 | top 4 | top 5 |
| 16 | 56.89 | 73.15 | 79.26 | 82.69 | 84.12 | 45.25 | 50.42 | 56.39 | 58.01 | 61.94 |
| 32 | 66.18 | 77.15 | 82.69 | 86.15 | 90.02 | 48.26 | 52.67 | 59.99 | 63.29 | 65.78 |
| 64 | 64.18 | 77.15 | 82.69 | 86.15 | 89.15 | 45.01 | 50.29 | 56.36 | 59.21 | 60.99 |
| 128 | 57.14 | 62.19 | 68.17 | 71.69 | 74.15 | 43.21 | 48.91 | 54.89 | 56.97 | 61.19 |
| 256 | 52.99 | 61.05 | 66.18 | 67.17 | 71.09 | 41.36 | 48.05 | 54.17 | 55.3 | 60.15 |

After adjusting the parameters of HMM using validation data, we tested the features in test data. Results in test dataset using different features are detailed in Table IV. The qualitative recognition results of middle zone using five features for both scripts are shown in Fig.20. Note that, few word images are not recognized by some features. Some examples of middle zone recognition results with PHOG feature taking top N choices are shown in Fig.21. These choices are next refined with upper and lower zone modifiers to obtain the correct results. The recognition accuracy using PHOG features according to word length is shown in Table V. It is noticed that when the number of characters in word increases the recognition accuracy improves. Fig. 22 shows the recognition performance of character and words in HMM-based middle zone recognition.

**Table III: Performance accuracy in tabular format for different state numbers.**

| | | **Bangla** | | | | | **Devanagari** | | | | |
|---|---|---|---|---|---|---|---|---|---|---|---|
| | State # | top 1 | top 2 | top 3 | top 4 | top 5 | top 1 | top 2 | top 3 | top 4 | top 5 |
| **LGH** | 6 | 69.58 | 76.25 | 77.59 | 82.36 | 83.69 | 70.28 | 77.14 | 78.48 | 83.69 | 84.19 |
| | 7 | 74.36 | 81.59 | 83.29 | 87.28 | 88.36 | 75.18 | 81.48 | 84.26 | 88.47 | 89.65 |
| | 8 | 77.09 | 86.61 | 88.25 | 90.35 | 91.21 | 78.77 | 87.49 | 89.17 | 92.15 | 92.45 |
| | 9 | 75.34 | 82.14 | 84.57 | 87.99 | 89.69 | 76.18 | 83.69 | 84.88 | 88.92 | 89.69 |
| | | **Bangla** | | | | | **Devanagari** | | | | |
| | State # | top 1 | top 2 | top 3 | top 4 | top 5 | top 1 | top 2 | top 3 | top 4 | top 5 |
| **PHOG** | 6 | 70.29 | 77.69 | 79.58 | 83.91 | 84.29 | 71.36 | 78.39 | 80.19 | 84.39 | 86.36 |
| | 7 | 75.36 | 82.15 | 83.99 | 88.17 | 89.59 | 76.98 | 83.67 | 84.78 | 89.65 | 90.69 |
| | 8 | 78.69 | 87.89 | 89.15 | 91.28 | 92.98 | 82.69 | 90.25 | 92.45 | 92.78 | 94.59 |
| | 9 | 75.97 | 83.64 | 85.45 | 88.47 | 90.21 | 77.24 | 85.25 | 86.69 | 90.14 | 91.39 |

| | PHOG | LGH | G-PHOG | GABOR | M-B |
|---|---|---|---|---|---|
| সম্মান | ✓সামলে | ✓সামলে | ✓সামলে | ✓সামলে | ✗মাহলে |
| আসল | ✓আসল | ✓আসল | ✓আসল | ✗আমল | ✓আসল |
| বাঝবাব | ✓বাঝবাব | ✗কাবব | ✗বাঝয়া | ✓বাঝবাব | ✗কাববাব |
| আপনাব | ✓আপনাব | ✗আপন | ✓আপনাব | ✗পাবচয | ✓আপনাব |
| বলেছিলেন | ✗বলপবক | ✓বলোছলেন | ✗কাহল | ✗পাডযা | ✗বালল |

(a)

| | PHOG | LGH | G-PHOG | GABOR | M-B |
|---|---|---|---|---|---|
| কম | ✓কম | ✓কম | ✓কম | ✗কভ | ✓কম |
| পশ্রা | ✗পসশা | ✗পসশা | ✗পসশা | ✗পসরা | ✗পসরা |
| সধার | ✗বধ | ✗বধ | ✓সধার | ✗বধ | ✗বধ |
| মনারজন | ✓মনারজন | ✓মনারজন | ✗মনায়া | ✗মনায়া | ✓মনারজন |
| জড় | ✓জড | ✓জড | ✓জড | ✓জড | ✓জড |

(b)

**Fig.20:** Few examples of middle zone recognition results using different features indicating correct (by tick) and incorrect (by cross) labels for (a) Bangla and (b) Devanagari scripts.

| Image | আবাব | কলকাতা | বনাব | জাম | সমথন |
|---|---|---|---|---|---|
| Choice 1 | ✓আবাব | ✗বনলতা | ✗কবব | ✓জাম | ✗সমান |
| Choice 2 | ✗তাবাব | ✗কমলতা | ✓বলাব | ✗জামা | ✗সময |
| Choice 3 | ✗অবাক | ✓কলকাতা | ✗কলম | ✗জন | ✗মহান |
| Choice 4 | ✗আকাব | ✗পলক | ✗কমাব | ✗জল | ✗সামথ |
| Choice 5 | ✗তাব | ✗কানন | ✗বলা | ✗হকাব | ✓সমথন |

(a)

| Image | जटने | आज | जगहा | समाधना | लगाम |
|---|---|---|---|---|---|
| Choice 1 | ✓जटत | ✗आपच | ✓जगह | ✗साधना | ✗लगय |
| Choice 2 | ✗जमल | ✗आपल | ✗जगहा | ✗साध | ✓लगाय |
| Choice 3 | ✗जात | ✓आज | ✗चमाना | ✗साधन | ✗लगा |
| Choice 4 | ✗जपज | ✗आन | ✗जहा | ✗सवध | ✗लगात |
| Choice 5 | ✗जगल | ✗आप | ✗जबका | ✓समाधना | ✗लाय |

(b)

**Fig.21. Recognition results of middle zone components of (a) Bangla and (b) Devanagari scripts considering Top N (=5) choices**

**Table IV: Middle zone recognition result with different features. Top 1 and Top 5 choices are shown here.**

| Script | Feature | Top-1 | Top-5 |
|---|---|---|---|
| Bangla | LGH | 77.21 | 91.07 |
| | PHOG | 79.17 | 92.89 |
| | G-PHOG | 65.34 | 88.71 |
| | GABOR | 48.21 | 70.23 |
| | Marti-Bunke | 45.25 | 64.72 |
| Devanagari | LGH | 79.29 | 92.61 |
| | PHOG | 82.11 | 94.51 |
| | G-PHOG | 66.27 | 89.62 |
| | GABOR | 48.47 | 71.82 |
| | Marti-Bunke | 47.51 | 65.87 |

**Table V: Middle zone recognition result with word image of different length**

| Script | Accuracy with word length | | | | | |
|---|---|---|---|---|---|---|
| | 2 | 3 | 4 | 5 | 6 | 7 |
| Bangla | 71.19 | 76.31 | 79.36 | 81.25 | 85.21 | 93.14 |
| Devanagari | 74.65 | 78.19 | 84.21 | 86.78 | 89.79 | 95.45 |

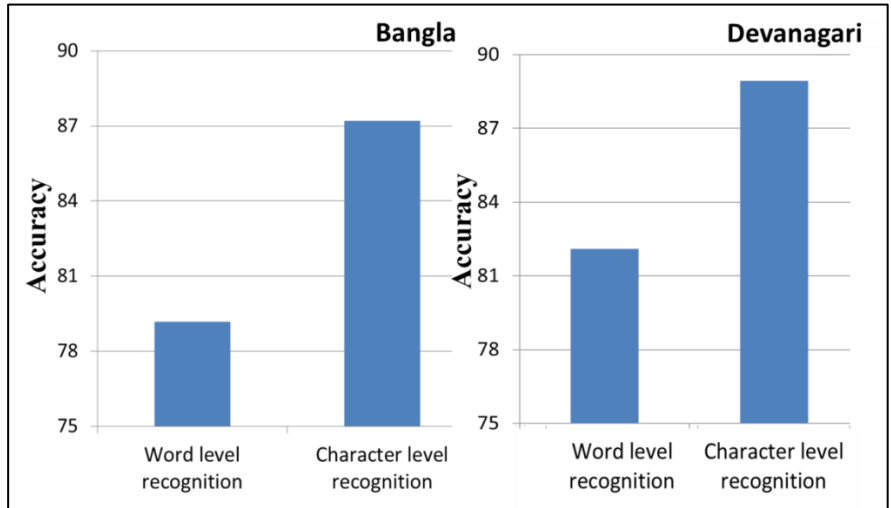

**Fig.22: Recognition performance of word and character level in middle zoned word image**

## 5.2. Upper and Lower Zone Recognition Results

For modifier recognition, we have collected a total of 1,723 upper zone modifiers and 1,437 lower zone modifiers from the training dataset for Bangla and a total of 1,656 upper zone modifiers and 1,351 lower zone modifiers for Devanagari script. To check the performance we have considered 500 modifiers for testing in each of these zones. Details of the data and performance analysis are shown in Table VI. The qualitative results using SVM are shown in Fig.23.

**Table VI: SVM classification result of the upper & lower zone modifiers**

| Script | Modifiers | Training data | Testing data | Accuracy (%) | |
|---|---|---|---|---|---|
| | | | | Top 1 | Top 2 |
| Bangla | Upper zone | 1,223 | 500 | 87.66 | 97.23 |
| | Lower zone | 937 | 500 | 84.07 | 95.15 |
| Devanagari | Upper zone | 1,156 | 500 | 85.95 | 96.52 |
| | Lower zone | 851 | 500 | 91.94 | 98.14 |

**Fig.23:** Some examples of modifier classification by SVM with indication of correct (by tick) & incorrect (by cross) one.

## 5.3. Full Word Recognition Result Combining Zone-wise Results

After getting zone-wise results from three zones, they are combined to form the full word. The combination is performed according to the mapping function $F$ as discussed in Section 4.2.3. The mapping function $F$ is used to join middle zone components with its counter-part from upper and lower zones using $C_i=F(C_{M\_i},C_{U\_i},C_{U\_i+1},C_{U\_i-1},C_{L\_i},C_{L\_i+1},C_{L\_i-1})$. Next, the lexicon of full word is searched using Levenshtein distance to provide the best result for the given component lists. Some examples of full word recognition results are shown in Fig.24. It is to be noted that these words have different skew and slant angles. Also, the upper and lower modifiers make the word recognition difficult. With our proposed system these words are recognized perfectly. The details of recognition performance at full word level are shown in Table VII. In Bangla we have achieved accuracy of 83.39% and 92.89% with top 1 and top 5 choices, respectively. 84.24% and 94.51% accuracy with top 1 and top 5 choices has been achieved in Devanagari script. Table VIII details the improvement in each step using proposed approach of word recognition.

**Fig.24:** Qualitative results of full word recognition (a) Bangla, (b) Devanagari

**Table VII: Accuracy of full word recognition using PHOG feature in middle zone**

| Script | Recognition Accuracy | | | | |
|---|---|---|---|---|---|
| | Top 1 | Top 2 | Top 3 | Top 4 | Top 5 |
| Bangla | 83.39 | 87.75 | 89.67 | 91.41 | 92.89 |
| Devanagari | 84.24 | 89.14 | 91.47 | 92.18 | 94.51 |

**Table VIII: Comparative study of successive improvement of full word recognition accuracy with implementation of newly proposed method (Here $\Delta I_{UZ}$ is improvement in upper zone segmentation using Water Reservoir principle, $\Delta I_{LZ}$ is improvement in lower zone segmentation using Shape matching algorithm, $\Delta I_C$ is improvement by combination of zone results using Refined character alignment.)**

| Method | Bangla | | Devanagari | |
|---|---|---|---|---|
| | Top -1 | Top -5 | Top-1 | Top-5 |
| Zone Segmentation Approach [35] | 80.21 | 90.87 | 81.31 | 91.94 |
| Zone Segmentation Approach + $\Delta I_{UZ}$ | 81.77 | 91.25 | 82.64 | 93.11 |
| Zone Segmentation Approach + $\Delta I_{UZ}$ + $\Delta I_{LZ}$ | 82.89 | 92.34 | 83.78 | 94.05 |
| Zone Segmentation Approach + $\Delta I_{UZ}$ + $\Delta I_{LZ}$ + $\Delta I_C$ | 83.39 | 92.82 | 84.24 | 94.51 |

## 5.4. Zone Segmentation Analysis

To understand the effectiveness of our zone segmentation strategies for complex writing styles, we have categorized the results of middle zone segmentation into three types, representing progressively errors in segmentation. InType-1, the zone segmentation is proper, Type-2 contains upto 10% errors, and finally in Type-3, when more than 10% of segmentation errors occur. Some images of different types are provided in Fig.25. Table IX shows the zone separation performance according to 3 different Types. We noticed that, Type-2 includes small segmentation errors in middle zone (e.g., a part of headline is missed, a small portion of upper/lower modifier is not segmented properly, etc.). We noted that these small errors do not affect much in our character recognition using SVM and HMM. Most of these words affecting with Type-2 error are recognized properly. In Type-3, if the middle zone recognition is not correct in first choice, we obtained correct recognition result in one of the N-best hypotheses. Next, when the middle zone characters are associated with upper and lower zone modifiers, we obtained better results.

It is to be noted that zone segmentation in Devanagari script was performed better using proposed approach. Presence of '*Matra*' is more horizontally straight in Devanagari script than Bangla script. Hence, zone segmentation of Devanagari script provided better accuracy than Bangla script.

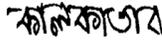

Fig.25: Some examples of zone segmentation using proposed approach from Type1, 2 and 3.

**Table IX: Zone segmentation performance using proposed approach.**

| Script | Type-1 | Type-2 | Type-3 |
|---|---|---|---|
| Bangla | 64.84% | 26.14% | 9.02% |
| Devanagari | 82.92% | 14.18% | 2.90 % |

Most of the segmentation errors in our approach are due to following. (a) The *Matra* detection approach may sometimes fail when $R_1$ and $R_2$ satisfy the *Matra* detection criteria but none of them provide the correct guess of the approximate headline row. This happens only when *Matra* is completely omitted while writing and the *Matra* rows are detected wrong (see Fig.26(a)).(b) The segmentation of lower zone modifiers is not perfect always. The lower zone modifier such as ও are sometimes mistakenly identified in the lower part of letter (see Fig.26(b)). (c) Some of the word images are very difficult to understand even manually. Due to cursive writings a part of the character is overlapped with other characters and hence our system could not recognize them (see Fig.26(c)).

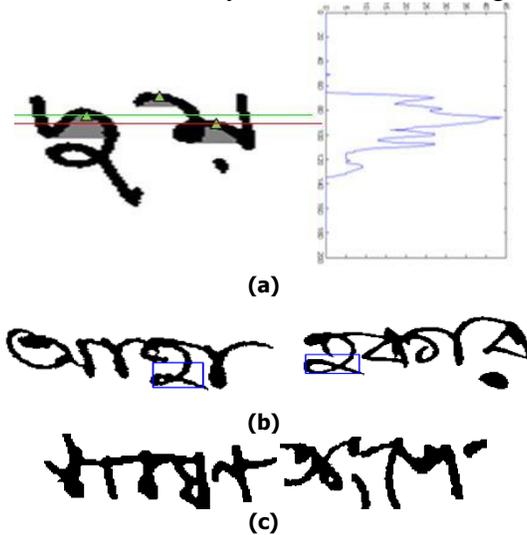

**Fig.26:(a)** Red line depicts the $R_1$ shown alongside with horizontal projection profile of the word. Green line denotes $R_2$ that is found using water resevoir concept.**(b)** Lower part of a letter is mistakenly identified as ও modifier. **(c)** Error in recognition due to too much bad handwriting.

### 5.5. Comparison of results

We have compared our result with the traditional full word recognition without using any zone segmentation. Fig. 27 shows the performance comparison when we applied without-zone segmentation based approach in our dataset. Sliding window features extracted using PHOG has been applied into HMM for recognition. The parameters are tuned to obtain the best results from the dataset. We obtained maximum43.21% (Bangla) and 42.65% (Devanagari) accuracy without zone segmentation. Whereas with our zone segmentation-based system we obtained 83.39% and 84.24% accuracy in Bangla and Devanagari, respectively. Thus, the effectiveness of our proposed zone segmentation-based approach can easily be justified. To compare with other systems, some results from existing systems are mentioned here. Though these results are not directly comparable as they are tested in different datasets we report some of them to have an idea. Shaw et al.[18] used combination of features for Devanagari word recognition using SVM based holistic approach. 81.14% and 84.02% accuracies were obtained with 100 and 50 word class problems respectively. Pal et al.[6] used Modified Quadratic Discriminant Function (MQDF) classifier based dynamic programming for Indic city name recognition. They obtained 94.08% and 90.16% accuracies in Bangla and Devanagari scripts with 84 and 117 words respectively.

Fig.28 compares the recognition results with our earlier approach [19] where zone segmentation approach used was not robust. In this present system, using 'Shape Matching' algorithm the segmentation of lower zone has been improved. Some qualitative results are shown in Fig.29. In Fig 29(b), we show zone segmentation results using method [19] which includes some artefacts in middle zone. Whereas, with proposed method, the middle zone segmentation is improved (See Fig.29(c).

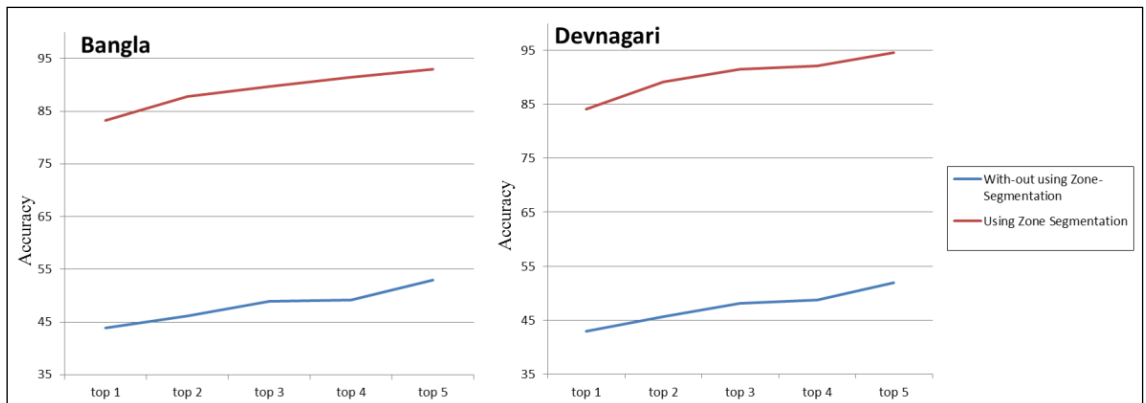

**Fig.27: Comparison of full word recognition result using zone-segmentation and with-out Zone-Segmentation**

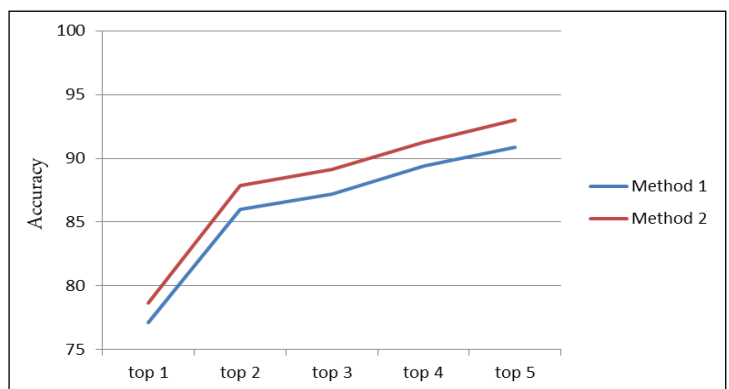

**Fig.28: Comparison of middle zone recognition result(Bangla) for two methods implemented for lower zone segmentation. Method 1 is due to the work of [19] and method 2 is the proposed system in this paper.**

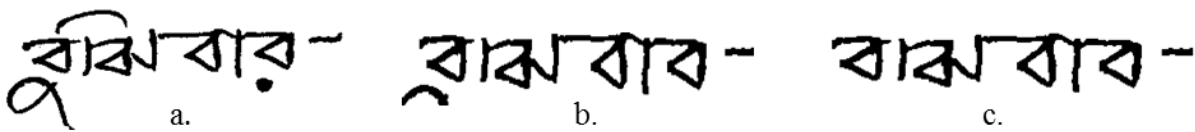

**Fig.29: a. Example of a word image, b. Zone segmentation using previous method [21] shows some artefacts in Middle zone. c. Zone segmentation using proposed method corrects such error and improved the recognition.**

### 5.6. Influence of Lexicon-Size in Recognition

Finally, an evaluation of the word recognition system was carried out on different lexicon sizes such as 1K, 2K, and 5K. The words in lexicon are collected from different sources like newspaper, online web-pages, etc. Table X details the recognition results of full words in both scripts. The numbers of unique words in our dataset are 1547 and 1957 for Bangla and Devanagari. Please note that, with increasing the lexicon size, the recognition performance gets reduced. With 5k lexicon size we achieved 70.47% and 71.30% accuracy in Bangla and Devanagari, respectively.Some examples of sentence recognition in Bangla and Devanagari scripts are shown in Fig.30. It is to be noted that the proposed system were able to recognize most of the words in each sentences.

**Table X: Full word recognition results with different lexicon size (1K, 2K, 5K)**

| Script | 1K | | 2K | | 5K | |
|---|---|---|---|---|---|---|
| | Top 1 | Top 5 | Top 1 | Top 5 | Top 1 | Top 5 |
| Bangla | 85.49 | 94.21 | 83.19 | 92.45 | 70.47 | 75.15 |
| Devanagari | 86.14 | 94.95 | 84.04 | 94.47 | 71.30 | 77.19 |

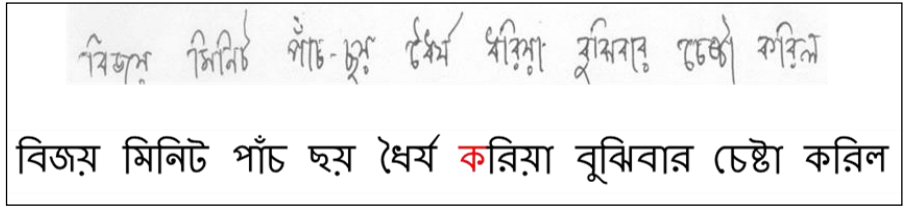

(a)

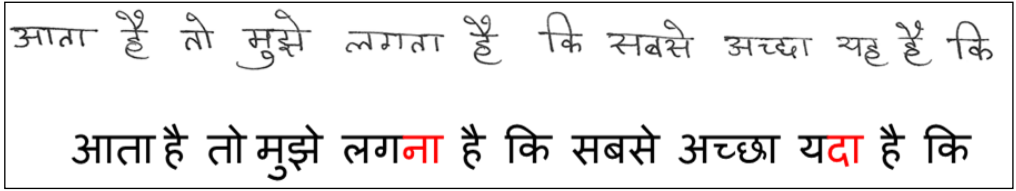

(b)

**Fig.30: Full sentence recognition result (a) Bangla (b) Devanagari (wrong recognition is indicated by red mark)**

## 5.7. Experiment with Noisy Images

We have tested our approach with the words added with synthetic noises. The words are degraded with Gaussian noise of different noise levels (10%, 20% and 30%). For recognition of noisy words, our approach is as follows. In the Binarized word image we apply a Gaussian smoothing technique to remove some of the noises. Next, an algorithm due to Roy et al. [32] is applied to join the broken contours in the image. If there is a small broken part in the component this algorithm can join the broken part and our proposed method work well. To get an idea of such word recognition results, some word images and corresponding results are shown in Fig.31. Here, the word images are added with 20% Gaussian noise. Quantitative results with noisy images obtained by10%,20%,30% Gaussian noise are shown in Fig.32.The performance dropped by approximately 9% for 20% noise in both datasets.

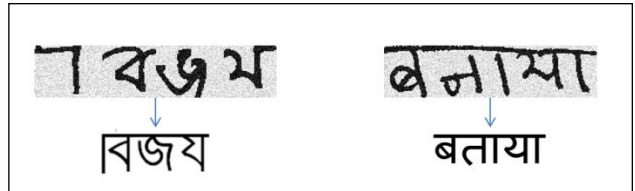

**Fig.31:Middle zone of word images corresponding results are shown. The images were added with Gaussian noises.**

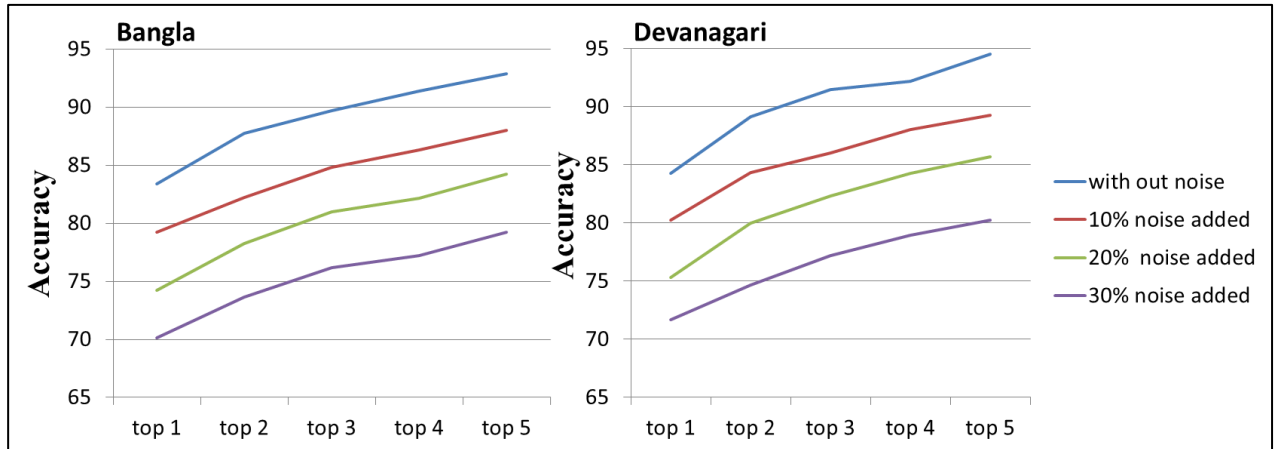
Fig.32: Comparison of full word recognition result for normal and noisy images.

## 6. Conclusion

In this paper, we have proposed a novel approach of Indic handwritten word recognition. A complete architecture is provided to use HMM-based segmentation free technique for efficient recognition. The proposed system segments a word into 3 zones and zone-wise recognize is made. As the proposed approach segments the zones of words, the training data entailed for character modelling is very less. Combining PHOG feature with HMM followed by segmentation helps to recognize the words in zone-wise. Since, the modifiers in upper and lower zones are distinct; SVM classifier is used for the identification of the modifiers of these two zones. Middle zone is recognized by HMM. Finally, zone-wise results are combined together to obtain the final word recognition rate. Different experiments are conducted to evaluate the efficiency and effectiveness of the proposed method. Our experiments have shown that zone based segmentation does have stronger recognition capability. So proper "zone segmentation" plays an important role to reduce the number of possible combination in character set. To conclude, we observe that zone-wise recognition is very important in Bangla/Devanagari scripts and such zone-wise idea could offer new insights into several other similar type scripts like Assamese, Gurumukhi etc.

## Acknowledgement

The authors would like to thank the anonymous reviewers for their constructive comments and suggestions to improve the quality of the paper.

## Reference

[1] L. Koerich, R. Sabourin and C. Y. Suen, "Recognition and verification of unconstrained handwritten words", IEEE Trans. Pattern Analysis and Machine Intelligence, vol.27, pp. 1509–1522, 2005.
[2] C. L. Liu and M. Koga and H. Fujisawa, "Lexicon driven segmentation and recognition of handwritten character strings for Japanese address reading", IEEE Trans. Pattern Analysis and Machine Intelligence, vol.24, pp.1425-1437, 2002.
[3] T. Su, "Chinese Handwriting Recognition: An Algorithmic Perspective", Springer, 2013.

[4] V. Märgner and H.E. Abed, "Arabic handwriting recognition competition", International Conf. on Document Analysis and Recognition, vol.2, pp. 1274-1278, 2007.

[5] U. Bhattacharya and B. B. Chaudhuri, "Handwritten Numeral Databases of Indian Scripts and Multistage Recognition of Mixed Numerals". IEEE Trans. Pattern Analysis and Machine Intelligence, vol.31, pp. 444-457, 2009.

[6] U. Pal, R. Roy and F. Kimura, "Multi-lingual City Name Recognition for Indian Postal Automation", International Conference on Frontiers in Handwriting Recognition, pp.169-173, 2012.

[7] T. Bhowmik, U. Roy and S. K. Parui, "Lexicon Reduction Technique for Bangla Handwritten Word Recognition", Document Analysis Systems, pp. 195-199, 2012

[8] Wikipedia reference, http://en.wikipedia.org/wiki/Bengali_alphabet.

[9] B. B. Chaudhuri and U. Pal, "A complete printed Bangla OCR system", Pattern Recognition, vol. 31(5), pp. 531-549, 1998.

[10] U. Pal and B. B. Chaudhuri, "Indian script character recognition: A survey", Pattern Recognition, vol.37, pp. 1887-1899, 2004.

[11] A. Hasnat, S. M. Habib and M. Khan. "A high performance domain specific OCR for Bangla script", Int. Joint Conf. on Computer, Information, and Systems Sciences, and Engineering (CISSE), 2007.

[12] R. Jayadevan, S.R. Kohle, P.M. Patil and U. Pal, "Offline recognition of Devanagari script: A survey". IEEE Trans. Systems, Man, and Cybernetics, Part C: Applications and Reviews, vol.41, No. 6, pp. 782-796, 2011.

[13] A.-L. Bianne-Bernard, F. Menasri, R. A.-H. Mohamad, C. Mokbel, C. Kermorvant, and L. Likforman-Sulem, "Dynamic and contextual information in HMM modeling for handwritten word recognition", IEEE Trans. Pattern Analysis and Machine Intelligence, vol. 33, pp. 2066-2080, 2011.

[14] M. Y. Chen, A. Kundu and J. Zhou, "Off-line handwritten word recognition using a Hidden Markov Model type stochastic network", IEEE Trans. Pattern Analysis and Machine Intelligence, vol.16, pp.481-496, 1994.

[15] R. Plamondon and S. N. Srihari, "On-line and off-line handwritten recognition: A comprehensive survey", IEEE Trans. Pattern Analysis and Machine Intelligence, vol.22, pp.62-84, 2000.

[16] S. Madhvanath and V. Govindaraju, "The role of holistic paradigms in handwritten word recognition", IEEE Trans. Pattern Analysis and Machine Intelligence, vol.23, pp.149-164, 2001.

[17] T. Plötz and G. A. Fink, "Markov models for offline handwriting recognition: a survey", International Journal on Document Analysis and Recognition, vol. 12(4), pp. 269-298, 2009.

[18] B. Shaw, U. Bhattacharya and S. K. Parui, "Combination of Features for Efficient Recognition of Offline Handwritten Devanagari Words", International Conference on Frontiers in Handwriting Recognition, pp. 240-245, 2014.

[19] P. P. Roy, P. Dey, S. Roy, U. Pal andF. Kimura, "A Novel Approach of Handwritten Text Recognition using HMM", International Conference on Frontiers in Handwriting Recognition, pp.661-666, 2014.

[20] P. P. Roy, U. Pal and J. Lladós, "Morphology Based Handwritten Line Segmentation Using Foreground and Background Information", International Conference on Frontiers in Handwriting Recognition, pp. 241-246, 2008.

[21] U. Pal, P. P. Roy, N. Tripathy and J. Lladós. "Multi-Oriented Bangla and Devnagari Text Recognition". Pattern Recognition, vol.43, pp. 4124-4136, 2010

[22] E. Kavallieratou, N. Fakotakis, and G. Kokkinakis, "Slant estimation algorithm for OCRsystem" Pattern Recognition, vol.34, pp. 2515–2522, 2001.

[23] C.-C. Chang and C.-J. Lin, "LIBSVM : a library for support vector machines", ACM Transactions on Intelligent Systems and Technology, pp. 1-27, 2011.

[24] T.-F. Wu, C.-J. Lin, and R. C. Weng, "Probability Estimates for Multi-class Classification by Pairwise Coupling", Journal of Machine Learning Research, pp.975-1005, 2004.

[25] N. Dalal and B. Triggs, "Histograms of Oriented Gradients for Human Detection", Computer Vision and Pattern Recognition, pp. 886-893, 2005

[26] Y. Bai, L. Guo, L. Jin and Q. Huang, "A novel feature extraction method using Pyramid Histogram of Orientation Gradients for smile recognition". International Conf. on Image Processing, pp.3305-3308, 2009.

[27] J. R. Serrano and F. Perronnin, "Handwritten word-spotting using Hidden Markov Models and universal vocabularies", International Conference on Pattern Recognition,vol.42, pp.2106-2116,2009.

[28] J. Chen, H. Cao, R. Prasad, A. Bhadwaj, and P. Natarajan, "Gabor features for offline Arabic handwriting recognition", Document Analysis Systems, pp.53–58,2010.

[29] H. Cao, R. Prasad and P. Natarajan, "Handwritten and Typewritten Text Identification and Recognition Using Hidden Markov Models", International Conf. on Document Analysis and Recognition, pp. 744-748, 2011.

[30] U.-V. Marti and H. Bunke, "Using a statistical language model to improve the performance of an HMM-based cursive handwriting recognition system," International Journal on Pattern Recognition and Artificial Intelligence, vol. 15, pp. 65–90, 2001.

[31] S. Young et al. The HTK Book, Version 3.4, 2006
[32] K. Roy, U. Pal and B. B. Chaudhuri, "A system for joining and recognition of broken Bangla numerals for Indian postal automation", In Proceedings of 4th Indian Conference on Computer Vision, Graphics and Image Processing, pp. 641–646, 2004.
[33] V. Vapnik, "The Nature of Statistical Learning Theory", Springer Verlang, 1995
[34] V. Levenshtein, "Binary codes capable of correcting deletions, insertions, and reversals", Soviet Physics Doklady vol. 10 (8), pp. 707–710, 1966.
[35] A. K. Bhunia, A. Das, P. P. Roy and U. Pal "A Comparative Study of Features for Handwritten Bangla Text Recognition", Proc. in International Conf. on Document Analysis and Recognition, pp. 636-640, 2015.
[36] http://www.iitr.ac.in/media/facspace/proy.fcs/IndicWord.rar